\documentclass[runningheads,a4paper,reqno]{llncs}
\pdfoutput=1
\usepackage[utf8]{inputenc}
\usepackage[T1]{fontenc}
\usepackage{comment}
\usepackage{llncsstuffthm}

\usepackage{bm}
\usepackage[leqno]{amsmath}
\usepackage{booktabs}
\usepackage{multirow}
\usepackage{xspace}
\usepackage{enumitem}
\usepackage{array}
\usepackage{tikz}
\usetikzlibrary{matrix,snakes,arrows,shapes,positioning,arrows.meta,shapes.geometric}
\usepackage{pgfplots}
\pgfplotsset{compat=1.17}
\usepgfplotslibrary{fillbetween}
\usepackage{forest}
\forestset{
  nice empty nodes/.style={
    for tree={calign=fixed edge angles},
    delay={where content={}{shape=coordinate,for siblings={anchor=north}}{}}
  },
}
\usepackage{latexsym}
\usepackage{wrapfig}
\usepackage{rotating}
\usepackage[hidelinks]{hyperref}
\usepackage{plabels}

\newcommand{\nhphantom}[1]{\sbox0{#1}\hspace*{-\the\wd0}}

\newcommand{\f}[1]{\mathsf{#1}}
\newcommand{\g}[1]{\mathit{#1}}
\mathchardef\hyph="2D
\renewcommand{\P}{\f{P}}
\renewcommand{\c}{\f{c}}
\renewcommand{\i}{\f{c}}
\renewcommand{\c}{\f{i}}
\renewcommand{\i}{\f{i}}
\newcommand{\D}{\f{D}}
\renewcommand{\d}{\f{D}}

\newcommand{\Dterms}{\mbox{D-terms}\xspace}
\newcommand{\Dterm}{\mbox{D-term}\xspace}
\newcommand{\DTerms}{\mbox{D-Terms}\xspace}
\newcommand{\DTerm}{\mbox{D-Term}\xspace}

\newcommand{\dterms}{\mbox{D-terms}\xspace}
\newcommand{\dterm}{\mbox{D-term}\xspace}

\newcommand{\tsize}{\f{t\hyph size}}
\newcommand{\csize}{\f{c\hyph size}}
\newcommand{\height}{\f{height}}
\newcommand{\scsize}{\f{sc\hyph size}}

\newcommand{\supterm}{\mathrel{\rhd}}
\newcommand{\suptermq}{\mathrel{\unrhd}}

\newcommand\subsumedBy{\mathrel{\ooalign{$\geq$\cr
      \hidewidth\raise.225ex\hbox{$\cdot\mkern7.0mu$}\cr}}}
\newcommand\subsumes{\mathrel{\ooalign{$\leq$\cr
      \hidewidth\raise.225ex\hbox{$\cdot\mkern2.0mu$}\cr}}}
\newcommand\strictlySubsumedBy{\mathrel{\ooalign{$\geq$\cr
      \hidewidth\raise.225ex\hbox{$\cdot\mkern7.0mu$}\cr}}}
\newcommand\variant{\mathrel{\ooalign{$=$\cr
      \hidewidth\raise.7ex\hbox{$\cdot\mkern4.5mu$}\cr}}}

\newcommand{\replace}[3]{#1{[}#2 \mapsto #3{]}}

\newcommand{\UCN}[1]{\forall #1}
\newcommand{\UCM}[1]{\forall\mspace{2mu} #1}
\newcommand{\UC}[1]{\forall\, #1}

\newcommand{\DC}{\mathcal{D}} 

\newcommand{\gtrc}{\mathrel{>_{\mathrm{c}}}}
\newcommand{\geqc}{\mathrel{\geq_{\mathrm{c}}}}

\newcommand{\m}[1]{\mathit{#1}}

\newcommand{\vars}{\m{\mathcal{V}\hspace{-0.11em}ar}}
\newcommand{\dprim}{\m{\mathcal{P}\hspace{-0.05em}rim}}
\newcommand{\dom}{\m{\mathcal{D}\hspace{-0.08em}om}}

\newcommand{\vrng}{\m{\mathcal{VR}\hspace{-0.02em}ng}}
\newcommand{\pos}{\m{\mathcal{P}\!os}}
\newcommand{\leafpos}{\m{\mathcal{L}\hspace{-0.05em}eaf\mathcal{P}\!os}}
\newcommand{\innerpos}{\m{\mathcal{I}\hspace{-0.05em}nner\mathcal{P}\!os}}
\newcommand{\posvar}{\m{\mathcal{P}\!os}\mathcal{V}ar}

\newcommand{\mgu}{\f{mgu}}
\newcommand{\emptypos}{\epsilon}
\newcommand{\pairing}{\f{pairing}}

\newcommand{\sshift}[1]{\f{shift}_{#1}}

\newcommand{\mf}[1]{\mathit{#1}}
\newcommand{\mgt}{\mf{Mgt}}
\newcommand{\ipt}{\mf{Ipt}}

\newcommand{\Det}{\text{\textit{Det}}\xspace}
\newcommand{\Luk}{\text{\textit{{\L}ukasiewicz}}\xspace}
\newcommand{\Syll}{\text{\textit{Syll}}\xspace}
\newcommand{\Peirce}{\text{\textit{Peirce}}\xspace}
\newcommand{\Simp}{\text{\textit{Simp}}\xspace}
\newcommand{\PSYLL}{\text{\textit{{\L}DS}}\xspace}

\newcommand{\name}[1]{\emph{#1}}
\newcommand{\defname}[1]{\emph{#1}}
\newcommand{\entails}{\models}

\newcommand{\imp}{\rightarrow}

\newcommand{\eqdef}{\; 
\raisebox{-0.1ex}[0mm]{$ \stackrel{\raisebox{-0.2ex}{\tiny 
      \textnormal{def}}}{=} $}\; }

\newcommand{\xeqdef}{\eqdef} 
\newcommand{\eeqdef}{\eqdef} 
\newcommand{\vy}{y}
\newcommand{\vx}[2]{x^{#1}_{#2}}
\newcommand{\daxioms}{\alpha}

\newcommand{\xconn}[1]{{\tiny \textbf{#1}}}

\newcommand{\tl}{1.7cm}
\newcommand{\tll}{0.85cm}
\newcommand{\tlll}{0.425}
\newcommand{\tllll}{0.25cm}

\newcommand{\mereq}{\hspace{0.15em}=\hspace{0.15em}}
\newcommand{\n}{\mathrm{n}}

\newcommand{\mer}[1]{#1}

\newcolumntype{L}[1]{>{\raggedright\let\newline\\\arraybackslash\hspace{0pt}}p{#1}}
\newcolumntype{C}[1]{>{\centering\let\newline\\\arraybackslash\hspace{0pt}}p{#1}}
\newcolumntype{R}[1]{>{\raggedleft\let\newline\\\arraybackslash\hspace{0pt}}p{#1}}
\newcolumntype{X}[1]{>{\raggedright\let\newline\\\arraybackslash\hspace{0pt}$}p{#1}<$}
\newcolumntype{Y}[1]{>{\centering\let\newline\\\arraybackslash\hspace{0pt}$}p{#1}<$}
\newcolumntype{Z}[1]{>{\raggedleft\let\newline\\\arraybackslash\hspace{0pt}$}p{#1}<$}

\newenvironment{arrayprf}
{\begin{array}{Z{2.5em}@{\hspace{1em}}X{32em}}}
{\end{array}}
\newenvironment{arrayprfeq}
{\begin{array}{Z{2.5em}@{\hspace{1em}}Y{1.5em}X{30.5em}}}
{\end{array}}

\newenvironment{VarDescription}[1]%
{\begin{list}{}{%
      \settowidth{\labelwidth}{\textit{#1:}}%
      \setlength{\leftmargin}{\labelwidth}\addtolength{\leftmargin}{\labelsep}}}%
{\end{list}}

\newenvironment{AttDescription}
{\begin{VarDescription}{DH$_R$}}
{\end{VarDescription}}

{\begin{list}
    {}
    {%
      \setlength{\topsep}{1ex}%
      }}
{\end{list}}

\newcommand{\iv}[2]{{[\![#1,\!#2]\!]}}

\newcommand{\sall}{\sigma}
\newcommand{\ssubold}{\tau}
\newcommand{\ssubnew}{\tau^{\prime}}
\newcommand{\NT}{\overline{T}}

\newcommand{\simpn}{\f{simp\hyph n}}

\newcommand{\Lukasiewicz}{{\L}u\-ka\-sie\-wicz\xspace}

\newcommand{\ProverN}{\name{Prover9}\xspace}
\newcommand{\KRHyper}{\name{KRHyper}\xspace}
\newcommand{\CMProver}{\name{CMProver}\xspace}
\newcommand{\OEIS}{\name{OEIS}\xspace}

\newcommand{\trskip}[1]{\ldots}

\newcommand{\XC}{C\xspace}

\newcommand{\PMER}{\textsf{MER}\xspace}

\newcommand{\MGT}{MGT\xspace}
\newcommand{\IPT}{IPT\xspace}

\newcommand{\IPTs}{IPTs\xspace}

\newcommand{\IF}{\textbf{IF}\xspace}

\makeatletter

\newcommand\boldpar{\@startsection{paragraph}{4}{\z@}%
  {-0.5\p@ \@plus -0.2\p@ \@minus -0.5\p@}%
  {-0.5em \@plus -0.22em \@minus -0.1em}%
  {\normalfont\normalsize\bfseries\boldmath}}
\makeatother

\begin{document}

\setlength{\abovedisplayskip}{0.6ex plus0.2ex}%
\setlength{\belowdisplayskip}{0.6ex plus0.2ex}%
\setlength{\abovedisplayshortskip}{0pt}%
\setlength{\belowdisplayshortskip}{0pt}%

\mainmatter  

\title{Learning from \Lukasiewicz and Meredith: Investigations into Proof
  Structures\\(Extended Version)\\[-12pt]}

\titlerunning{Investigations into Proof Structures}

\author{\mbox{Christoph Wernhard\inst{1} \and Wolfgang Bibel\inst{2}}}

\institute{
Berlin, Germany
\email{info@christophwernhard.com}
\and
  Technical University Darmstadt, Germany
  \email{bibel@gmx.net}}

\newcommand{\nocontentsline}[3]{}
\newcommand{\tocless}[1]{\bgroup\let\addcontentsline=\nocontentsline{#1}\egroup}
\tocless{\maketitle}

\vspace{-0.7cm}

\begin{abstract}
  The material presented in this paper contributes to establishing a basis
  deemed essential for substantial progress in Automated Deduction. It
  identifies and studies global features in selected problems and their proofs
  which offer the potential of guiding proof search in a more direct way. The
  studied problems are of the wide-spread form of ``axiom(s) and rule(s) imply
  goal(s)''. The features include the well-known concept of lemmas. For their
  elaboration both human and automated proofs of selected theorems are taken
  into a close comparative consideration. The study at the same time accounts
  for a coherent and comprehensive formal reconstruction of historical work by
  \Lukasiewicz, Meredith and others.  First experiments resulting from the
  study indicate novel ways of lemma generation to supplement automated
  first-order provers of various families, strengthening in particular their
  ability to find short proofs.\linebreak
\end{abstract}

\vspace{-33pt}
\section{Introduction}
\vspace{-5pt}
\label{sec:intro}

Research in Automated Deduction, also known as Automated Theorem Proving
(ATP), has resulted in systems with a remarkable performance. Yet, deep
mathematical theorems or otherwise complex statements still withstand any of
the systems' attempts to find a proof. The present paper is motivated by the
thesis that the reason for the failure in more complex problems lies in the
local orientedness of all our current methods for proof search like resolution
or connection calculi in use.

In order to find out more global features for directing proof search we start
out here to study the structures of proofs for complex formulas in some detail
and compare human proofs with those generated by systems. Complex formulas of
this kind have been considered by \Lukasiewicz in~\cite{luk:1948}. They are
complex in the sense that current systems require tens of thousands or even
millions of search steps for finding a proof if any, although the length of
the formulas is very short indeed. How come that \Lukasiewicz found proofs
for those formulas although he could never carry out more than, say, a few
hundred search steps by hand? Which global strategies guided him in finding those
proofs? Could we discover such strategies from the formulas' global features?

By studying the proofs in detail we hope to come closer to answers to those
questions.  Thus it is proofs, rather than just formulas or clauses as usually
in ATP, which is in the focus of our study. In a sense we are aiming at an
ATP-oriented part of Proof Theory, a discipline usually pursued in Logic yet
under quite different aspects. This meta-level perspective has rarely been
taken in ATP for which reason we cannot rely on the existing conceptual basis
of ATP but have to build an extensive conceptual basis for such a study more or
less from scratch.

This investigation thus analyzes structures of, and operations on, proofs for
formulas of the form ``axiom(s) and rule(s) imply goal(s)''. It renders
condensed detachment, a logical rule historically introduced in the course of
studying these complex proofs, as a restricted form of the Connection Method
(CM) in ATP. All this is pursued with the goal of enhancing proof search in ATP
in mind.
As noted, our investigations are guided by a close inspection into proofs by
{\L}ukasiewicz and Meredith. In fact, the work presented here amounts at the
same time to a very detailed reconstruction of those historical proofs.

The rest of the paper is organized as follows: In Sect.~\ref{sec-background}
we introduce the problem and a formal human proof that guides our
investigations and compare different views on proof structures.  We then
reconstruct in Sect.~\ref{sec-basis} the historical method of \name{condensed
  detachment} in a novel way as a restricted variation of the CM where proof
structures are represented as terms.  This is followed in Sect.~\ref{sec-red}
by results on reducing the size of such proof terms for application in proof
shortening and restricting the proof search
space. Section~\ref{sec-properties} presents a detailed feature table for the
investigated human proof, and Sect.~\ref{sec-experiments} shows first
experiments where the features and new techniques are used to supplement the
inputs of ATP systems with lemmas. Section~\ref{sec-conclusion} concludes the
paper.  Supplementary technical material including proofs is provided in
Appendix~\ref{sec-appendix}. Data and tools to reproduce the experiments
are available at
\href{http://cs.christophwernhard.com/cd}{\texttt{http://cs.christophwernhard.com/cd}}.

\vspace{-10pt}
\section{Relating Formal Human Proofs with ATP Proofs}
\label{sec-background}
\vspace{-5pt}

\begin{figure}[t]
  \centering
  \begin{minipage}{27em}
    \vspace{1cm}
    $\P\i(\i(\i pq,r),\i(\i rp,\i sp))\land (\P x\wedge \P\i xy\rightarrow
    \P y)\rightarrow \P\i(\i\f{pq},\i(\i\f{qr},\i\f{pr}))$
       {
         \setlength{\unitlength}{0.28mm}
         \begin{picture}(0.00,0.00)(-15.00,3.10)
           \qbezier(-6.00,29.00)(54.00,47.00)(114.00,29.00)
           \put(52,29){\scriptsize\bf 5}
           \qbezier(-9.00,32.00)(66.00,56.00)(140.00,30.00)
           \put(66,45){\scriptsize\bf 4}
           \qbezier(118.00,10.00)(152.00,-10.00)(186.00,11.00)
           \put(151,-10){\scriptsize\bf 3}
           \qbezier(144.00,12.00)(163.00,0.00)(182.00,12.00)
           \put(162,7){\scriptsize\bf 2}
           \qbezier(186.00,28.00)(204.00,42.00)(222.00,28.00)
           \put(202,36){\scriptsize\bf 1}
         \end{picture}
       }
  \end{minipage}
  \vspace{0.2cm}
  \caption{\PSYLL along with its five unifiable connections.}
  \label{fig-ConnT}
  \vspace{-0.5cm}
\end{figure}
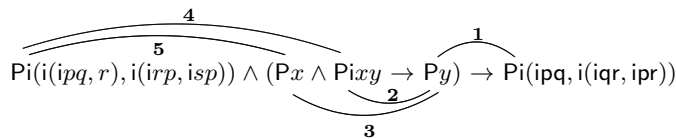

In 1948 Jan \Lukasiewicz published a formal proof of the completeness of his
shortest single axiom for the implicational fragment (\textbf{IF}), that is,
classical propositional logic with implication as the only logic operator
\cite{luk:1948}.  In his notation the implication $p \imp q$ is written as
$\g{Cpq}$.  Following Frank Pfenning~\cite{pfenning:single:1988} we formalize
\textbf{IF} on the meta-level in the first-order setting of modern ATP with a
single unary predicate $\P$ to be interpreted as something like ``provable''
and represent the propositional formulas by terms using the binary function
symbol $\c$ for implication. We will be concerned with the following
formulas.
\begin{center}\small
  \begin{tabular}{l@{\hspace{1em}}l@{\hspace{1em}}l}
   Nickname  \cite{prior:logicians:1956}\cite[p.~319]{prior:formal:logic:1962} & \Lukasiewicz's notation & First-order
   representation\\\midrule
  \Simp & $\g{CpCqp}$ & $\forall pq\, \P(\c(p,\c qp))$\\
  \Peirce & $\g{CCCpqpp}$ & $\forall pq\, \P(\c(\c(\c pq),p),p)$\\
  \Syll & $\g{CCpqCCqrCpr}$ & $\forall pqr\, \P(\c(\c pq,\c(\c qr,\c pr)))$\\
  \name{Syll Simp} & $\g{CCCpqrCqr}$ & $\forall pqr\,
\P\c(\c(\c pq,r),\c qr)$\\
  \Luk & $\g{CCCpqrCCrpCsp}$ & $\forall pqrs\, \P(\c(\c(\c pq,r),\c(\c rp,\c sp)))$
  \end{tabular}
\end{center}
\IF can be axiomatized by the set of the three axioms \Simp, \Peirce and
\Syll, known as \name{Tarski-Bernays Axioms}.  Alfred Tarski in 1925 raised
the problem to characterize \IF by a single axiom and solved it with very long
axioms, which led to a search for the \emph{shortest} single axiom, which was
found with the axiom nicknamed after him in 1936 by \Lukasiewicz
\cite{luk:1948}.  In 1948 he published his derivation that \Luk entails the
three Tarski-Bernays Axioms, expressed formally by the \emph{method of
  substitution and detachment}.
Detachment is also familiar as \name{modus ponens}. \Lukasiewicz's proof
involves 34 applications of detachment. Among the Tarski-Bernays axioms \Syll
is by far the most challenging to prove, hence his proof centers around the
proof of \Syll, with \Peirce and \Simp spinning off as side results.
Carew A. Meredith presented in \cite{meredith:notes:1963} a ``very slight
abridgement'' of \Lukasiewicz's proof, expressed in his framework of
\name{condensed detachment} \cite{prior:logicians:1956}, where the performed
substitutions are no longer explicitly presented but implicitly assumed
through unification. Meredith's proof involves only 33 applications of
detachment.
In our first-order setting, detachment can be modeled with the following
meta-level axiom.
\[\Det\; \eqdef\; \forall xy\, (\P x \land \P\c xy \imp \P y).\]
In~\Det the atom $\P x$ is called the \name{minor premise}, $\P\c xy$ the
\name{major premise}, and $\P y$ the \name{conclusion}.
Let us now focus on the following particular formula.
\[\PSYLL\; \eqdef\; \Luk \land \Det \imp \Syll.\]
``Problem~\PSYLL'' is then the problem of determining the validity of the
first order formula~\PSYLL.
In view of the CM \cite{bibel:atp:1982,bibel:deduction:1993,bibel:otten:2020},
a formula is valid if there is a spanning and complementary set of connections
in it.  In Fig.~\ref{fig-ConnT} \PSYLL is presented again, nicknames
dereferenced and quantifiers omitted as usual in ATP, with the five unifiable
connections in it. Observe that $p,q,r,s$ on the left side of the main
implication are variables, while $\f{p},\f{q},\f{r}$ on the right side are
Skolem constants.
Any CM proof of \PSYLL consists of a number of instances of the five shown
connections. Meredith's proof, for example, corresponds to 491~instances of
\Det, each linked with three instances of its five incident connections.

\begin{figure}[t]
  \normalsize
  \renewcommand{\xconn}[1]{{\scriptsize \textbf{#1}}}

\noindent(a)\\[-14pt]
\noindent
\hspace*{-0.28cm}
\scalebox{0.722}{{
    \begin{tikzpicture}
      \input{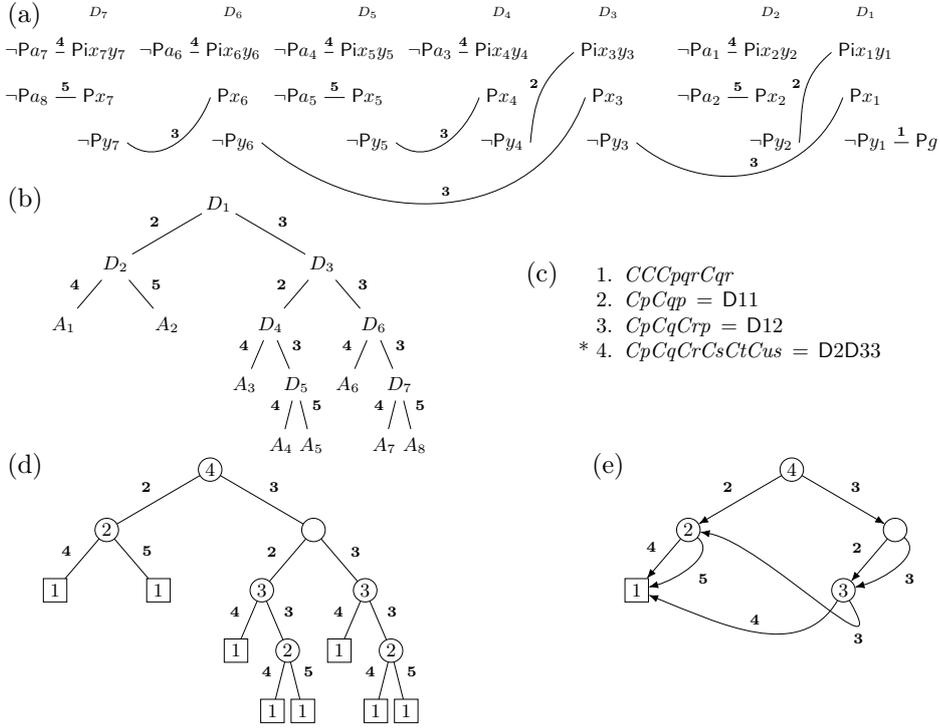}
    \end{tikzpicture}
}}

\vspace{-0.76cm}
  
\noindent(b)
\raisebox{0.2cm}{%
\scalebox{0.8}{%
\begin{tikzpicture}
  [baseline=(current bounding box.north),
    xdot/.style = {minimum size=11pt,
      inner sep=0pt, outer sep=2pt, on grid},
    xsqr/.style = {minimum size=11pt,
      inner sep=0pt, outer sep=2pt, on grid}]  
\node[xdot] (e) {\small $D_1$};
\node[xdot, below left=1cm and \tl of e] (1) {\small $D_2$};
\node[xdot, below right=1cm and \tl of e] (2) {\small $D_3$};
\node[xsqr, below left=1cm and \tll of 1] (11) {\small $A_1$};
\node[xsqr, below right=1cm and \tll of 1] (12) {\small $A_2$};
\node[xdot, below left=1cm and \tll of 2] (21) {\small $D_4$};
\node[xdot, below right=1cm and \tll of 2] (22) {\small $D_6$};
\node[xsqr, below left=1cm and \tlll of 21] (211) {\small $A_3$};
\node[xdot, below right=1cm and \tlll of 21] (212) {\small $D_5$};
\node[xsqr, below left=1cm and \tlll of 22] (221) {\small $A_6$};
\node[xdot, below right=1cm and \tlll of 22] (222) {\small $D_7$};
\node[xsqr, below left=1cm and \tllll of 212] (2121) {\small $A_4$};
\node[xsqr, below right=1cm and \tllll of 212] (2122) {\small $A_5$};
\node[xsqr, below left=1cm and \tllll of 222] (2221) {\small $A_7$};
\node[xsqr, below right=1cm and \tllll of 222] (2222) {\small $A_8$};
\draw[] (e) -- node [pos=0.5,above left] {\xconn{2}} (1);
\draw[] (e) -- node [pos=0.5,above right] {\xconn{3}} (2);
\draw[] (1) -- node [pos=0.6,above left] {\xconn{4}} (11);
\draw[] (1) -- node [pos=0.6,above right] {\xconn{5}} (12);
\draw[] (2) -- node [pos=0.6,above left] {\xconn{2}} (21);
\draw[] (2) -- node [pos=0.6,above right] {\xconn{3}} (22);
\draw[] (21) -- node [pos=0.6,above left] {\xconn{4}} (211);
\draw[] (21) -- node [pos=0.6,above right] {\xconn{3}} (212);
\draw[] (22) -- node [pos=0.6,above left] {\xconn{4}} (221);
\draw[] (22) -- node [pos=0.6,above right] {\xconn{3}} (222);
\draw[] (212) -- node [pos=0.6,above left] {\xconn{4}} (2121);
\draw[] (212) -- node [pos=0.6,above right] {\xconn{5}} (2122);
\draw[] (222) -- node [pos=0.6,above left] {\xconn{4}} (2221);
\draw[] (222) -- node [pos=0.6,above right] {\xconn{5}} (2222);
\end{tikzpicture}}}%
\hspace{35pt}%
\raisebox{-28pt}{%
\noindent(c)
\scalebox{0.8}{%
\begin{tabular}[t]{R{2em}@{\hspace{0.5em}}l}
 1. & $\g{CCCpqrCqr}$\\
 2. & $\g{CpCqp} \mereq \f{D}11$\\
 3. & $\g{CpCqCrp} \mereq \f{D}12$\\
 * 4. & $\g{CpCqCrCsCtCus} \mereq \f{D}2\f{D}33$\\
  \end{tabular}}}\\[-7pt]

\noindent(d)%
\raisebox{0.2cm}{%
\scalebox{0.8}{%
\begin{tikzpicture}
  [baseline=(current bounding box.north),
    dot/.style = {circle, minimum size=11pt,
      inner sep=0pt, outer sep=0pt, draw, on grid},
    sqr/.style = {regular polygon sides=4, minimum size=11pt,
      inner sep=0pt, outer sep=0pt, draw, on grid}]  
\node[dot] (e) {\small $4$};
\node[dot, below left=1cm and \tl of e] (1) {\small $2$};
\node[dot, below right=1cm and \tl of e] (2) {\small $$};
\node[sqr, below left=1cm and \tll of 1] (11) {\small $1$};
\node[sqr, below right=1cm and \tll of 1] (12) {\small $1$};
\node[dot, below left=1cm and \tll of 2] (21) {\small $3$};
\node[dot, below right=1cm and \tll of 2] (22) {\small $3$};
\node[sqr, below left=1cm and \tlll of 21] (211) {\small $1$};
\node[dot, below right=1cm and \tlll of 21] (212) {\small $2$};
\node[sqr, below left=1cm and \tlll of 22] (221) {\small $1$};
\node[dot, below right=1cm and \tlll of 22] (222) {\small $2$};
\node[sqr, below left=1cm and \tllll of 212] (2121) {\small $1$};
\node[sqr, below right=1cm and \tllll of 212] (2122) {\small $1$};
\node[sqr, below left=1cm and \tllll of 222] (2221) {\small $1$};
\node[sqr, below right=1cm and \tllll of 222] (2222) {\small $1$};
\draw[] (e) -- node [pos=0.5,above left] {\xconn{2}} (1);
\draw[] (e) -- node [pos=0.5,above right] {\xconn{3}} (2);
\draw[] (1) -- node [pos=0.6,above left] {\xconn{4}} (11);
\draw[] (1) -- node [pos=0.6,above right] {\xconn{5}} (12);
\draw[] (2) -- node [pos=0.6,above left] {\xconn{2}} (21);
\draw[] (2) -- node [pos=0.6,above right] {\xconn{3}} (22);
\draw[] (21) -- node [pos=0.6,above left] {\xconn{4}} (211);
\draw[] (21) -- node [pos=0.6,above right] {\xconn{3}} (212);
\draw[] (22) -- node [pos=0.6,above left] {\xconn{4}} (221);
\draw[] (22) -- node [pos=0.6,above right] {\xconn{3}} (222);
\draw[] (212) -- node [pos=0.6,above left] {\xconn{4}} (2121);
\draw[] (212) -- node [pos=0.6,above right] {\xconn{5}} (2122);
\draw[] (222) -- node [pos=0.6,above left] {\xconn{4}} (2221);
\draw[] (222) -- node [pos=0.6,above right] {\xconn{5}} (2222);
\end{tikzpicture}}}
\hspace{\fill}%
(e)%
\raisebox{0.2cm}{%
\scalebox{0.8}{%
\begin{tikzpicture}
  [baseline=(current bounding box.north),
    >={Latex[length=4.5pt]},
    dot/.style = {circle, minimum size=11pt,
      inner sep=0pt, outer sep=0pt, draw, on grid},
    sqr/.style = {regular polygon sides=4, minimum size=11pt,
      inner sep=0pt, outer sep=0pt, draw, on grid}]  
\node[dot] (e) {\small $4$};
\node[dot, below left=1cm and \tl of e] (1) {\small $2$};
\node[dot, below right=1cm and \tl of e] (2) {\small $$};
\node[sqr, below left=1cm and \tll of 1] (11) {\small $1$};
\node[dot, below left=1cm and \tll of 2] (21) {\small $3$};
\draw[->] (e) -- node [pos=0.5,above left] {\xconn{2}} (1);
\draw[->] (e) -- node [pos=0.5,above right] {\xconn{3}} (2);
\draw[->] (1) -- node [pos=0.5,above left] {\xconn{4}} (11);
\draw[->] (1) to [out=-40,in=15] node [pos=0.5,below right] {\xconn{5}} (11);
\draw[->] (2) -- node [pos=0.5,above left] {\xconn{2}} (21);
\draw[->] (2) to [out=-40,in=15] node [pos=0.5,below right] {\xconn{3}} (21);
\draw[->] (21) to [out=240,in=-22] node [pos=0.5,above] {\xconn{4}} (11);
\draw[->] (21) to [out=-60,in=-10,out looseness=1.5] node [pos=0.2,below] {\xconn{3}} (1);
\end{tikzpicture}}}
  \caption{A proof in different representations.}
  \label{fig-representations}
  \vspace{-0.6cm}
\end{figure}

Figure~\ref{fig-representations} compares different representations of a short
formal proof with the \Det meta axiom.  There is a single axiom, \name{Syll
  Simp}, and the theorem is $\forall pqrstu\, \P \c(p,\c(q,\c(r,\c(s,\c(t,\c
us)))))$.
Figure~\ref{fig-representations}a shows the structure of a CM proof. It
involves seven instances of \Det, shown in columns $D_1, \ldots, D_7$.  The
major premise $\P\c x_i y_i$ is displayed there on top of the minor
premise~$\P x_i$, and the (negated) conclusion $\lnot \P y_i$, where $x_i,
y_i$ are variables.  Instances of the axiom appear as literals $\lnot \P a_i$,
with $a_i$ a shorthand for the term $\c(\c(\c p_i q_i,r_i),\c q_i r_i)$.  The
rightmost literal~$\P g$ is a shorthand for the Skolemized theorem.
The clause instances are linked through edges representing connection
instances.  The edge labels identify the respective connections as in
Fig.~\ref{fig-ConnT}.  An actual connection proof is obtained by supplementing
this structure with a substitution under which all pairs of literals related
through a connection instance become complementary.

Figure~\ref{fig-representations}b represents the \emph{tree} implicit in the
CM proof. Its inner nodes correspond to the instances of \Det, and its leaf
nodes to the instances of the axiom. Edges appear ordered to the effect that
those originating in a major premise of \Det are directed to the left and
those from a minor premise to the right. The goal clause~$\P g$ is dropped.
The resulting tree is a \emph{full binary tree}, i.e., a binary tree where
each node has 0 or 2 children.  We observe that the ordering of the children
makes the connection labeling redundant as it directly corresponds to the tree
structure.

Figure~\ref{fig-representations}c presents the proof in Meredith's notation.
Each line shows a formula, line~1 the axiom and lines 2--4 derived formulas,
with proofs annotated in the last column.  Proofs are written as terms in
Polish notation with the binary function symbol $\f{D}$ for \name{detachment}
where the subproofs of the major and minor premise are supplied as first and
second, resp., argument.  Formula~4, for example, is obtained as
conclusion of \Det applied to formula~2 as major premise and as minor premise
another formula that is not made explicit in the presentation, namely the
conclusion of \Det applied to formula~3 as both, major and minor, premises.
An asterisk marks the goal theorem.

Figure~\ref{fig-representations}d is like Fig.~\ref{fig-representations}b, but
with a different labeling: Node labels now refer to the line in
Fig.~\ref{fig-representations}c that corresponds to the subproof rooted at the
node. The blank node represents the mentioned subproof of the formula that is
not made explicit in Fig.~\ref{fig-representations}b.  An inner node
represents a condensed detachment step applied to the subproof of the major
premise (left child) and minor premise (right child).

Figure~\ref{fig-representations}e shows a DAG (directed acyclic graph)
representation of~Figure~\ref{fig-representations}d.  It is the unique
maximally factored DAG representation of the tree, i.e., it has no multiple
occurrences of the same subtree.  Each of the four proof line labels of
Fig.~\ref{fig-representations}c appears exactly once in the DAG.

\begin{wrapfigure}[19]{R}{0.561\textwidth}
  \centering\small
  \vspace{-3pt}
  \hspace{-5pt}
  \begin{tabular}{rR{1.29em}@{\hspace{0.2em}}l}
   & 1. & $\g{CCCpqrCCrpCsp}$\\
    &  2. & $\g{CCCpqpCrp} \mereq \f{DDD}1\f{D}111\n$\\ 
    &  3. & $\g{CCCpqrCqr}  \mereq  \f{DDD}1\f{D}1\f{D}121\n$\\
    &  4. & $\g{CpCCpqCrq}  \mereq  \f{D}31$\\
    &  5. & $\g{CCCpqCrsCCCqtsCrs}  \mereq  \f{DDD}1\f{D}1\f{D}1\f{D}141\n$\\
    &  6. & $\g{CCCpqCrsCCpsCrs}  \mereq  \f{D}51$\\
    &  7. & $\g{CCpCqrCCCpsrCqr}  \mereq  \f{D}64$\\
    &  8. & $\g{CCCCCpqrtCspCCrpCsp}  \mereq  \f{D}71$\\
    &  9. & $\g{CCpqCpq}  \mereq  \f{D}83$\\
    & 10. & $\g{CCCCrpCtpCCCpqrsCuCCCpqrs}  \mereq  \f{D}18$\\
    & 11. & $\g{CCCCpqrCsqCCCqtsCpq}  \mereq  \f{DD}10.10.\n$\\
    & 12. & $\g{CCCCpqrCsqCCCqtpCsq}  \mereq  \f{D}5.11$\\
    & 13. & $\g{CCCCpqrsCCsqCpq}  \mereq  \f{D}12.6$\\
    & 14. & $\g{CCCpqrCCrpp}  \mereq  \f{D}12.9$\\
    & 15. & $\g{CpCCpqq}  \mereq  \f{D}3.14$\\
    & 16. & $\g{CCpqCCCprqq}  \mereq  \f{D}6.15$\\
    *$\!\!$ & 17. & $\g{CCpqCCqrCpr}  \mereq  \f{DD}13.\f{D}16.16.13$\\
    *$\!\!$ & 18. & $\g{CCCpqpp}  \mereq  \f{D}14.9$\\
    *$\!\!$ & 19. & $\g{CpCqp}  \mereq  \f{D}33$\\
  \end{tabular}
  \vspace{-14pt}
  \caption{Proof \PMER, Meredith's refinement \cite{meredith:notes:1963} of
    \Lukasiewicz's proof \cite{luk:1948}.}
  \label{fig-proof-mer}
\end{wrapfigure}

We conclude this introductory section with reproducing Meredith's refinement
of \Lukasiewicz's completeness proof in Fig.~\ref{fig-proof-mer}, taken from
\cite{meredith:notes:1963}.  Since we will often refer to this proof, we call
it \PMER.  There is a single axiom~(1), which is \Luk. The proven theorems are
\Syll (17), \Peirce (18) and \Simp (19).  In addition to line numbers also the
symbol $\n$ appears in some of the proof terms. Its meaning will be explained
later on in the context of Def.~\ref{def-simp-n}. For now, we can read $\n$ just as
``1''.  Dots are used in the Polish notation to disambiguate numeric
identifiers with more than a single digit.

\vspace{-9pt}
\section{Condensed Detachment and a Formal Basis}
\vspace{-8pt}
\enlargethispage{2pt}
\label{sec-basis}

Following \cite{bunder:cd:1995}, the idea of condensed detachment can be
described as follows: Given premises \mbox{$F \imp G$} and $H$, we can
conclude $G'$, where $G'$ is the most general result that can be obtained by
using a substitution instance $H'$ as minor premise with the substitution
instance $F' \imp G'$ as major premise in modus ponens. Condensed detachment
was introduced by Meredith in the mid-1950s as an evolution of the earlier
\name{method of substitution and detachment}, where the involved substitutions
were explicitly given.  The original presentations of condensed detachment are
informal by means of examples
\cite{prior:logicians:1956,lemmon:meredith:purestrict:1957,prior:formal:logic:1962,meredith:memoriam:1977},
formal specifications have been given later
\cite{kalman:cd:1983,hindley:meredith:cd:1990,bunder:cd:1995}.
In ATP, the rendering of condensed detachment by hyperresolution with the
clausal form of axiom \Det is so far the prevalent view.  As overviewed in
\cite{mccune:wos:cd:1992,ulrich:legacy:2001}, many of the early successes of
ATP were based on condensed detachment.  Starting from the hyperresolution
view, structural aspects of condensed detachment have been considered by
Robert Veroff \cite{veroff:shortest:2001} with the use of term representations
of proofs and \emph{linked} resolution.  Results of ATP systems on deriving
the Tarski-Bernays axioms from \Luk are reported in
\cite{pfenning:single:1988,wos:contributes:1990,roo:parallel:1992,mccune:wos:cd:1992,fitelson:missing:2001}.
Our goal in this section is to provide a formal framework that makes the
achievements of condensed detachment accessible from a modern ATP view.  In
particular, the incorporation of unification, the interplay of nested
structures with explicitly and implicitly associated formulas, sharing of
structures through lemmas, and the availability of proof structures as terms.

\boldpar{Notation.}  Most of our notation follows common practice
\cite{dershowitz:notations:1991} such that we only provide some reminding
hints here: $s \subsumedBy t$ expresses that $t$ subsumes $s$ and $s \suptermq
t$ that $t$ is a subterm of $s$.  A \defname{position} is a sequence of
positive integers that specifies the occurrence of a subterm in a term as a
path in Dewey decimal notation starting from the root of term. The set of all
positions of a term $s$ is denoted by $\pos(s)$.  For example,
$\pos(\f{f}(x,\f{g}(y))) = \{\emptypos, 1, 2, 2.1\}$. For $p \in \pos(s)$,
$s|_p$ denotes the \defname{subterm of $s$ at position $p$}, and $s[t]_p$ the
term obtained from $s$ by replacing the subterm occurrence at position $p$
with term~$t$.

In addition, we make use of a few special symbols and conventions: The set of
positions $p \in \pos(s)$ such that $s|_p$ is a variable or a constant is
denoted by $\leafpos(s)$ and the set of positions $p \in \pos(s)$ such that
$s|_p$ is a compound term by~$\innerpos(s)$.  We use the postfix notation for
the application of a substitution~$\sigma$ also for sets~$M$ of pairs of
terms: $M\sigma$ stands for $\{\{s\sigma, t\sigma\} \mid \{s, t\} \in M\}$.
For terms $s,t,u$, the expression $\replace{s}{t}{u}$ denotes $s$ after
simultaneously replacing all occurrences of $t$ with $u$.  The
\defname{height} $\height(s)$ of a term~$s$ is, viewing the term as a tree,
the number of edges of the longest downward path from the root to a leaf. In
the literature it also called \name{depth} of the term.  If $F$ is a formula,
then $\UCN{F}$ denotes the universal closure of $F$.

\vspace{-10pt}
\subsection{Proof Structures: \DTerms, Tree Size and Compacted Size}
\label{sec-struct-base}
\vspace{-4pt}

In this section we consider only the purely structural aspects of condensed
detachment proofs.  Emphasis is on a twofold view on the proof structure, as a
tree and as a DAG (directed acyclic graph), which factorizes multiple
occurrences of the same subtree.  Both representation forms are useful: the
compacted DAG form captures that lemmas can be repeatedly used in a proof,
whereas the tree form facilitates to specify properties in an inductive
manner.  We call the tree representation of proofs by terms with the binary
function symbol~$\d$ \name{\dterms}.

\vspace{-5pt}
\begin{defn}
  \label{def:d-term}
  (i)~We assume a distinguished set of symbols called
  \defname{primitive \dterms}. (ii)~A \defname{\dterm} is inductively
  specified as follows: (1.)~Any primitive \dterm is a \dterm.  (2.)~If $d_1$
  and $d_2$ are \dterms, then $\d(d_1,d_2)$ is a \dterm.  (iii)~The set of
  primitive \dterms occurring in a \dterm~$d$ is denoted by $\dprim(d)$.
  (iv)~The set of all \dterms that are not primitive is denoted by $\DC$.
\end{defn}
\vspace{-5pt}

\noindent
A \dterm is a full binary tree (i.e, a binary tree in which every node has
either 0 or 2 children), where the leaves are labeled with symbols, i.e.,
primitive \dterms.  An example \dterm is
\begin{equation}
  \label{eq-examp-dterm}
  d\; \eqdef\; \d(\d(1,1),\d(\d(1,\d(1,1)),\d(1,\d(1,1)))),
\end{equation}
which represents the structure of the proof shown in
Fig.~\ref{fig-representations} and can be visualized by the full binary tree
of Fig.~\ref{fig-representations}d after removing all labels with exception of
the leaf labels.
The proof annotations in Fig.~\ref{fig-representations}c and
Fig.~\ref{fig-proof-mer} are \dterms written in Polish notation. The
expression $\f{D}2\f{D}33$ in line~4 of Fig.~\ref{fig-representations}, for
example, stands for the \dterm $\d(2,\d(3,3))$. $\dprim(\d(2,\d(3,3))) =
\{2,3\}$.

A finite tree and, more generally, a finite set of finite trees can be
represented as DAG, where each node in the DAG corresponds to a subtree of a
tree in the given set.  It is well known that there is a unique \name{minimal}
such DAG, which is maximally factored (it has no multiple occurrences of the
same subtree) or, equivalently, is minimal with respect to the number of
nodes, and, moreover, can be computed in linear time
\cite{downey:variations:1980}.  The number of nodes of the minimal DAG is the
number of distinct subtrees of the members of the set of trees.  There are two
useful notions of measuring the size of a \dterm, based directly on its tree
representation and based on its minimal DAG, respectively.
\begin{defn}
  (i)~The \defname{tree size} of a \dterm $d$, in symbols $\tsize(d)$, is the
  number of occurrences of the function symbol~$\D$ in $d$.  (ii)~The
  \defname{compacted size}\footnote{We took the notion of \name{compacted
      size} from \cite{flajolet:1990}.}  of a
  \dterm $d$ is defined as $\csize(d)\; \eqdef\; |\{e \in \DC \mid d \suptermq
  e \}|.$ (iii)~The \defname{compacted size} of a finite set $D$ of \dterms is
  defined as $\csize(D)\; \eqdef\; |\{e \in \DC \mid d \in D \text{ and } d
  \suptermq e \}|.$
\end{defn}

\vspace{-5pt}

\noindent
The tree size of a \dterm can equivalently be characterized as the number of
its inner nodes.  The compacted size of a \dterm is the number of its distinct
compound subterms. It can equivalently be characterized as the number of the
inner nodes of its minimal DAG.
As an example consider the \dterm $d$ defined in formula~(\ref{eq-examp-dterm}), whose
minimal DAG is shown in Fig.~\ref{fig-representations}e. The tree size of $d$
is $\tsize(d) = 7$ and the compacted size of $d$ is $\csize(d) = 4$,
corresponding to the cardinality of the set $\{e \in \DC \mid d \suptermq e
\}$ of compound subterms of~$d$, i.e., $\{\D(1,1),\; \D(1,\D(1,1)),\;
\D(\D(1,\D(1,1)),\D(1,\D(1,1))),\; d\}.$

The tabular presentation of proof~\PMER (Fig.~\ref{fig-proof-mer}) renders its
DAG structure as a mapping of the line numbers to the trees, i.e., \Dterms, in
the right column.\footnote{We only sketch the reading of Meredith's tabular
  notation as a mapping that specifies a DAG here.  It can be formally
  considered, for example, as a substitution or as a regular tree grammar. In
  \cite{cwwb-article} we use the concept of \name{compacted \Dterm} for this.}
The DAG represents a set of three trees corresponding to proofs of \Syll
(line~17), \Peirce (line~18) and \Simp (line~19), respectively. The compacted
size of the set of these three is~33, which can be determined by counting the
occurrences of $\f{D}$ in the right column. For the individual subproofs, the
compacted size can be determined by counting the occurrences of $\f{D}$ in
only those lines that can be reached via the mapping from the respective root,
and the tree size by counting the occurrences of $\f{D}$ after unfolding the
respective root according to the mapping.  The proof of \Syll in \PMER, for
example, has compacted size~31 and tree size~491.

As will be explicated in more detail below, each occurrence of the function
symbol~$\D$ in a \dterm corresponds to an instance of the meta-level axiom
$\Det$ in the represented proof. Hence the tree size measures the number of
instances of $\Det$ in the proof. Another view is that each occurrence of
$\f{D}$ in a \dterm corresponds to a condensed detachment step, without
re-using already proven lemmas.  The compacted size of a \dterm is the number
of its distinct compound subterms, corresponding to the view that the size of
the proof of a lemma is only counted once, even if it is used multiply in the
proof.  Tree size and compacted size of \dterms have been previously
identified as relevant proof size measures in \cite{veroff:shortest:2001},
called there \name{CDcount} and \name{length}, respectively.

\vspace{-14pt}
\subsection{Proof Structures, Formula Substitutions and Semantics}
\label{sec-subst-base}
\vspace{-6pt}
\enlargethispage{13pt}

We use a notion of \name{unifier} that applies to a set of pairs of terms, as
common in discussions based on the CM
\cite{bibel:atp:1982,eder:subst:1985,eder:relative:1992}.  Although a unifier
of a finite set of pairs~$\{\{s_1,t_1\},\ldots,\{s_n,t_n\}\}$ can be expressed
as unifier of the single pair $\{\f{f}(s_1,\ldots,s_n),
\f{f}(t_1,\ldots,t_n)\}$ of terms, the explicit definition for a set of pairs
is useful because such pairs naturally arise in the CM and the related proof
trees, \dterms, in condensed detachment.

\vspace{-5pt}
\begin{defn}
  \label{def-mgu-content}%
Let~$M$ be a set of pairs of terms and let $\sigma$ be a substitution.
(i)~$\sigma$ is said to be a \defname{unifier} of $M$ if for all $\{s,t\} \in
M$ it holds that $s\sigma = t\sigma$.  (ii)~$\sigma$ is called a \defname{most
  general unifier} of $M$ if $\sigma$ is a unifier of $M$ and for all
unifiers~$\sigma'$
of $M$ it holds that $\sigma' \subsumedBy \sigma$.
(iii)~$\sigma$ is called a \defname{clean most general unifier}
  of $M$ if it is a most general unifier of $M$ and, in addition, is
  idempotent and satisfies $\dom(\sigma) \cup \vrng(\sigma) \subseteq
  \vars(M)$.
\end{defn}

\vspace{-5pt}

\noindent
The additional properties required for \name{clean} most general unifiers do
not hold for all most general unifiers.\footnote{The inaccuracy observed by
  \cite{hindley:meredith:cd:1990} in early formalizations of condensed
  detachment based on the notion of \name{most general unifier} can be
  attributed to the failure of considering the requirement $\dom(\sigma) \cup
  \vrng(\sigma) \subseteq \vars(M)$ of the \name{clean} property.}  However,
the unification algorithms known from the literature produce \emph{clean} most
general unifiers \cite[Remark 4.2]{eder:subst:1985}. If a set of pairs of
terms has a unifier, then it has a most general unifier and, moreover, also a
\emph{clean} most general unifier.

\vspace{-5pt}
\begin{defn}
  \subdefinline{def-mgu}~If $M$ is a set of pairs of terms that has a unifier,
  then $\mgu(M)$ denotes some clean most general unifier of $M$.  $M$ is
  called \defname{unifiable} and $\mgu(M)$ is called \defname{defined} in this
  case, otherwise it is called \defname{undefined}.
  \subdefinline{def-mgu-partial-convention}~We make the convention that
  proposition, lemma and theorem statements implicitly assert their claims
  only for the case where occurrences of $\mgu$ in them are defined.
\end{defn}
\vspace{-5pt}

\noindent
Since we define $\mgu(M)$ as a \emph{clean} most general unifier, we are
permitted to make use of the assumption that it is idempotent and that all
variables occurring in its domain and range occur in $M$.
Convention~\ref{def-mgu-partial-convention} has the purpose to reduce clutter
in proposition, lemma and theorem statements.

The structural aspects of condensed detachment proofs represented by\linebreak \dterms,
i.e., full binary trees, will now be supplemented with associated formulas.
Condensed detachment proofs, similar to CM proofs, involve different instances
of the input formulas (viewed as quantifier free, e.g., clauses), which may be
considered as obtained in two steps: first, ``copies'', that is, variants with
fresh variables, of the input formulas are created; second a substitution is
applied to these copies.
Let us consider now the first step.  The framework of \dterms permits to give
the variables in the copies canonical designators with an index subscript that
\emph{identifies the position in the structure}, i.e., in the \dterm, or
tree.

\vspace{-5pt}
\begin{defn}
  For all positions~$p$ and positive integers $i$ let $\vx{i}{p}$ and $\vy_p$
  denote pairwise different variables.
\end{defn}
\vspace{-5pt}
\enlargethispage{8pt}

\noindent
Recall that positions are path specifiers.  For a given \dterm~$d$ and leaf
position $p$ of~$d$ the variables $\vx{i}{p}$ are for use in a formula
associated with~$p$ which is the copy of an axiom.  Different variables in the
copy are distinguished by the upper index~$i$. If $p$ is a non-leaf position
of~$d$, then~$\vy_p$ denotes the variable in the conclusion of the copy of
$\Det$ that is represented by~$p$. In addition, $\vy_p$ for leaf positions~$p$
may occur in the antecedents of the copies of $\Det$.  The following
equivalences, which hold for all positions $p$, justify this coupling of
positions and the variables $\vx{i}{p}, \vy_p$ for \Luk as an example of an
application axiom and for \Det.
\begin{eqnarray}
  \Luk & \;\equiv\; & \UC{
\P(\c(\c(\c(\vx{1}{p},\vx{2}{p}),\vx{3}{p}),
\c(\c(\vx{3}{p},\vx{1}{p}),\c(\vx{4}{p},\vx{1}{p}))))}.
  \label{eq-luk-canonical}\\
 \Det & \;\equiv\; & \UC{(\P(\c(\vy_{p.2},\vy_{p})) \land \P(\vy_{p.2})
   \imp \P(\vy_p))}. \label{eq-det-canonical}
\end{eqnarray}
Here the major premise of \Det appears to the left of the minor one, matching
the argument order of the $\d$ function symbol.  The following substitution
$\sshift{p}$ is a tool to systematically rename position-associated variables
while preserving the internal relationships between the index-referenced
positions.

\begin{defn}
  \label{def-sshift}%
  For all positions~$p$ define the substitution $\sshift{p}$ as follows:
  $\sshift{p}\linebreak \eqdef\; \{\vy_{q} \mapsto \vy_{p.q} \mid q \text{ is a
    position}\} \cup \{\vx{i}{q} \mapsto \vx{i}{p.q} \mid i \geq 1 \text{ and
  } q \text{ is a position}\}$.
\end{defn}

\vspace{-3pt}

\noindent
The application of $\sshift{p}$ to a term~$s$ effects that $p$ is prepended to
the position indexes of all the position-associated variables occurring
in~$s$.
The association of axioms with primitive \dterms is
represented by mappings which we call \name{axiom assignments}, defined as
follows.

\vspace{-3pt}

\begin{defn}
  An \defname{axiom assignment} $\daxioms$ is a mapping whose domain is a set
  of primitive \dterms and whose range is a set of terms whose variables are
  in $\{\vx{i}{\emptypos} \mid i \geq 1\}$.  We say that $\daxioms$ is
  \defname{for} a \dterm~$d$ if $\dom(\daxioms) \supseteq \dprim(d)$.
\end{defn}

\vspace{-3pt}

\noindent
We define a shorthand for a form of \Luk that is suitable for use as a range
element of axiom assignments. It is parameterized with a position~$p$.
  \begin{equation}\Luk_{p}\; \eqdef\;
  \c(\c(\c(\vx{1}{p},\vx{2}{p}),\vx{3}{p}),\c(\c(\vx{3}{p},\vx{1}{p}),\c(\vx{4}{p},\vx{1}{p}))).\end{equation}
The mapping $\{1 \mapsto \Luk_{\emptypos}\}$ is an axiom assignment for
all \dterms~$d$ with $\dprim(d) = \{1\}$.  In Meredith's proof presentation
the axiom assignment is represented by the steps with no trailing \dterm, such
as lines~1 in Fig.~\ref{fig-representations}c and~\ref{fig-proof-mer}.
The second step of obtaining the instances involved in a proof can be
performed by applying the most general unifier of a pair of terms that
constrain it.  The tree structure of \dterms permits to associate exactly one
such pair with each term position. Inner positions represent detachment steps
and leaf positions instances of an axiom according to a given axiom
assignment.  The following definition specifies these constraining pairs.

\vspace{-3pt}

\begin{defn}
  \label{def-pairing}
  Let $d$ be a \dterm and let $\daxioms$ be an axiom assignment for $d$.  For
  all positions $p \in \pos(d)$ define the pair of terms
  $\pairing_{\daxioms}(d, p)\ \eqdef\ \{\vy_p,\, \daxioms(d|_p)\sshift{p}\}\linebreak
  \text{ if } p \in \leafpos(d)$ and $\{\vy_{p.1},\, \c(\vy_{p.2}, \vy_p)\}
  \text{ if } p \in \innerpos(d)$.
\end{defn}

\vspace{-3pt}

\noindent
A unifier of the set of pairings of all positions of a \dterm~$d$ equates for
a leaf position~$p$ the variable $y_p$ with the value of the axiom
assignment~$\alpha$ for the primitive \dterm at $p$, after ``shifting''
variables by~$p$.  This ``shifting'' means that the position subscript
$\emptypos$ of the variables in the axiom argument term~$\alpha(d|_p)$ is
replaced by $p$, yielding a dedicated copy of the axiom argument term for the
leaf position~$p$.  For inner positions~$p$ the unifier equates $\vy_{p.1}$
and $\c(\vy_{p.2}, \vy_p)$, reflecting that the major premise of \Det is
proven by the left child of~$p$.

The substitution induced by the pairings associated with the positions of a
\dterm allow to associate a specific formula with each position of the \dterm,
called the \name{in-place theorem (\IPT)}.  The case where the position is the
top position~$\emptypos$ is distinguished as \name{most general theorem (\MGT)}.

\vspace{-3pt}

\begin{defn}
  \label{def-ipt-mgt}%
  For \dterms~$d$, positions $p \in \pos(d)$ and axiom assignments $\daxioms$
  for $d$ define the \defname{in-place theorem (\IPT) of}~$d$ \defname{at}~$p$
  \defname{for}~$\daxioms$, $\ipt_{\daxioms}(d,p)$, and the \defname{most
    general theorem (\MGT) of}~$d$ \defname{for}~$\daxioms$,
  $\mgt_{\daxioms}(d)$, as \subdefinline{def-ipt}~$\ipt_{\daxioms}(d,p)\;
  \eqdef$\linebreak $\P(y_{p}\mgu(\{\pairing_{\daxioms}(d, q) \mid q \in
  \pos(d)\})).$ \subdefinline{def-mgt}~$\mgt_{\daxioms}(d)\; \eqdef
  \ipt_{\daxioms}(d,\emptypos)$.
\end{defn}

\vspace{-3pt}
\enlargethispage{10pt}

\noindent
Since $\ipt$ and $\mgt$ are defined on the basis of $\mgu$, they are undefined
if the set of pairs of terms underlying the respective application of $\mgu$
is not unifiable. Hence, we apply the convention of
Def.~\ref{def-mgu-partial-convention} for $\mgu$ also to occurrences of
$\ipt$ and $\mgt$.  If $\ipt$ and $\mgt$ are defined, they both denote an atom whose
variables are constrained by the \name{clean} property of the underlying
application of $\mgu$.
The following proposition relates \IPT and \MGT with
respect to subsumption.
\begin{prop}
 \label{prop-ipt-subsumedby-mgt}
  For all \dterms~$d$, positions $p \in \pos(d)$ and axiom assignments
  $\daxioms$ for $d$ it holds that $\ipt_{\daxioms}(d,p) \subsumedBy
  \mgt_{\daxioms}(d|_p).$
\end{prop}

\vspace{-4pt}

\noindent
By Prop.~\ref{prop-ipt-subsumedby-mgt}, the \IPT at some
position~$p$ of a \dterm~$d$ is subsumed by the \MGT of the
subterm~$d|_p$ of $d$ rooted at position~$p$. An intuitive argument is that
the only constraints that determine the most general unifier underlying the
\MGT are induced by positions of $d|_p$, that is,
\emph{below}~$p$ (including~$p$ itself). In contrast, the most general unifier
underlying the \IPT is determined by \emph{all} positions of $d$.

The following lemma expresses the core relationships between a proof structure
(a \dterm), a proof substitution (accessed via the \IPT) and
semantic entailment of associated formulas.

\vspace{-4pt}

\begin{lem}
  \label{lem-core}
  Let $d$ be a \dterm and let $\daxioms$ be an axiom assignment for $d$. Then
  for all $p \in \pos(d)$ it holds that: \subpropinline{lem-core-leaf}~If $p
  \in \leafpos(d)$, then $\UCM{\P(\daxioms(d|_p))}\; \entails\;
  \ipt_{\daxioms}(d,p).$ \subpropinline{lem-core-inner}~If $p \in
  \innerpos(d)$, then $\Det \land \ipt_{\daxioms}(d,p.1) \land
  \ipt_{\daxioms}(d,p.2)\; \entails\; \ipt_{\daxioms}(d,p).$
\end{lem}

\vspace{-4pt}

\noindent
Based on this lemma, the following theorem shows how \name{Detachment}
together with the axioms in an axiom assignment entail the \MGT of a given
\dterm.

\vspace{-4pt}

\begin{thm}
  \label{thm-sem-mgt}
  Let $d$ be a \dterm and let $\daxioms$ be an axiom assignment for $d$.  Then
  $\Det \land \bigwedge_{p \in \leafpos(d)}
  \UCM{\P(\daxioms(d|_p))} \;\entails\; \UCN{\mgt_{\daxioms}(d)}.$
\end{thm}

\vspace{-4pt}

\noindent
Theorem~\ref{thm-sem-mgt} states that \Det together with the axioms referenced
in the proof, that is, the values of~$\alpha$ for the leaf nodes of~$d$
considered as universally closed atoms, entail the universal closure of the
\MGT of~$d$ for~$\alpha$.  The universal closure of the \MGT is the formula
exhibited in Meredith's proof notation in the lines with a trailing \dterm,
such as lines~2--19 in Fig.~\ref{fig-proof-mer}.

\vspace{-11pt}
\section{Reducing the Proof Size by Replacing Subproofs}
\label{sec-red}
\vspace{-7pt}
\enlargethispage{6.5pt}

The term view on proof trees suggests to shorten proofs by rewriting subterms,
that is, replacing occurrences of subproofs by other ones, with three main
aims: (1)~To shorten given proofs, with respect to the tree size or the
compacted size. (2)~To investigate given proofs whether they can be shortened
by certain rewritings or are closed under these. (3)~To develop notions of
redundancy for use in proof search.  A proof fragment constructed during
search may be rejected if it can be rewritten to a shorter one.

It is obvious that if a \dterm~$d'$ is obtained from a \dterm~$d$ by replacing
an occurrence of a subterm $e$ with a \dterm $e'$ such that $\tsize(e) \geq
\tsize(e')$, then also $\tsize(d) \geq \tsize(d')$.  Based on the following
ordering relations on \dterms, which we call \emph{compaction orderings}, an
analogy for reducing the \emph{compacted} size instead of the tree size can be
stated.

\vspace{-4pt}

\begin{defn}
  \label{def-c-orderings}%
  For \dterms $d,e$ define \subdefinline{def:geqc}~$d \geqc e\;\eqdef\; \{f
  \in \DC \mid d \supterm f\} \supseteq \{f \in \DC \mid e \supterm f\}.$
  \subdefinline{def:gtrc}~$d \gtrc e\; \eqdef\; d \geqc e \text{ and } e \not
  \geqc d.$
\end{defn}

\vspace{-4pt}

\noindent
The relations $d \geqc e$ and $d \gtrc e$ compare \dterms~$d$ and $e$ with
respect to the superset relationship of their sets of those strict subterms
that are compound terms.  For example, $\D(\D(\D(1,1),1),1) \gtrc
\D(1,\D(1,1))$ because $\{\D(1,1),\, \D(\D(1,1),1)\}\linebreak \supseteq
\{\D(1,1)\}$.

\vspace{-4pt}

\begin{thm}
  \label{thm-replace-csize-scsize}
  Let $d,d',e,e'$ be \mbox{\dterms} such that $e$ occurs in $d$, and $d' =
  \replace{d}{e}{e'}$. It holds that \subthminline{thm-replace-csize}~If $e
  \in \DC$ and $e \geqc e'$, then $\csize(d) \geq \csize(d')$.
  \subthminline{thm-replace-scsize}~If $e \gtrc e'$, then $\scsize(d) >
  \scsize(d')$, where, for all \dterms~$d$
  $\scsize(d) \eqdef \sum_{d \suptermq e} \csize(e)$.
\end{thm}
Theorem~\ref{thm-replace-csize} states that if $d'$ is the \dterm obtained
from $d$ by simultaneously replacing \emph{all} occurrences of a compound
\dterm~$e$ with a ``c-smaller'' \dterm~$e'$, i.e., $e \geqc e'$, then the
compacted size of $d'$ is less or equal to that of $d$.  As stated with the
supplementary Theorem~\ref{thm-replace-scsize}, the $\scsize$ is a measure that
strictly decreases under the strict precondition $e \gtrc e'$, which is useful
to ensure termination of rewriting.
The following proposition characterizes the number of \dterms that are smaller
than a given \dterm w.r.t the compaction ordering~$\geqc$.\hspace{-3pt}
\begin{prop}
  \label{prop-co-number}
  For all \dterms $d$ it holds that
  $|\{e \mid d \geqc e \text{ and } \dprim(e) \subseteq \dprim(d)\}|
  = (\csize(d) - 1 + |\dprim(d)|)^2 + |\dprim(d)|.$
\end{prop}
By Prop.~\ref{prop-co-number}, for a given \dterm~$d$, the number of
\dterms~$e$ that are smaller than~$d$ with respect to $\geqc$ is only
quadratically larger than the compacted size of~$d$ and hence also than the
tree size of~$d$. Hence techniques that inspect all these smaller \dterms for
a given \dterm can efficiently be used in practice.

According to Theorem~\ref{thm-sem-mgt}, a condensed detachment proof, i.e.,
a \dterm~$d$ and an axiom assignment~$\alpha$, proves the \MGT
of~$d$ for~$\alpha$ along with instances of the \MGT.  In
general, replacing subterms of~$d$ should yield a proof of at least these
theorems.  That is, a proof whose \MGT subsumes the original
one. Hence we are interested in identifying conditions that ensure that
subterm replacement steps yield proofs with a \MGT that
subsumes the \MGT before the replacement.  The following
theorems express such conditions.
\begin{thm}
  \label{thm-sr-ipt}
  Let $d, e$ be \dterms, let~$\daxioms$ be an axiom assignment for $d$ and for
  $e$, and let $p_1, \ldots, p_n$, where $n \geq 0$, be positions in $\pos(d)$
  such that for all $i,j \in \{1,\ldots,n\}$ with $i \neq j$ it holds that
  $p_i \not \leq p_j$.  If for all $i \in \{1, \ldots, n\}$ it holds that
  $\ipt_{\daxioms}(d, p_i) \subsumedBy \mgt_{\daxioms}(e)$, then
  $\mgt_{\daxioms}(d) \subsumedBy \mgt_{\daxioms}(d[e]_{p_1}[e]_{p_2}\ldots
  [e]_{p_n})$.
\end{thm}
Theorem~\ref{thm-sr-ipt} states that simultaneously replacing a number of
occurrences of possibly different subterms in a \dterm by the same subterm
with the property that its \MGT subsumes each of the \IPTs of the original
occurrences results in an overall \dterm whose MGT
subsumes that of the original overall \dterm.
The following theorem is similar, but restricted to the case of a single
replaced subterm occurrence and with a stronger precondition.  It follows from
Theorem~\ref{thm-sr-ipt} and Prop.~\ref{prop-ipt-subsumedby-mgt}.
\begin{thm}
  \label{thm-sr-mgt}
  Let $d, e$ be \dterms and let~$\daxioms$ be an axiom assignment for $d$ and
  for $e$.  For all positions $p \in \pos(d)$ it then holds that if
  $\mgt_{\daxioms}(d|_p) \subsumedBy \mgt_{\daxioms}(e)$, then
  $\mgt_{\daxioms}(d) \subsumedBy \mgt_{\daxioms}(d[e]_p)$.
\end{thm}
\emph{Simultaneous} replacements of subterm occurrences are essential for
reducing the compacted size of proofs according to
Theorem~\ref{thm-replace-csize-scsize}.  For replacements according to
Theorem~\ref{thm-sr-mgt} they can be achieved by successive replacements of
individual occurrences. In Theorem~\ref{thm-sr-ipt} simultaneous replacements
are explicitly considered because the replacement of one occurrence according
to this theorem can invalidate the preconditions for another occurrence.
Theorem~\ref{thm-sr-mgt} can be useful in practice because the precondition
$\mgt_{\daxioms}(d|_p) \subsumedBy \mgt_{\daxioms}(e)$ can be evaluated on the
basis of $\alpha$, $e$ and \emph{just the subterm} $d|_p$ of $d$, whereas
determining $\ipt_{\daxioms}(d,p)$ for Theorem~\ref{thm-sr-ipt} requires also
consideration of the \emph{context} of $p$ in~$d$.
Based on Theorems~\ref{thm-sr-ipt} and~\ref{thm-replace-csize-scsize} we
define the following notions of reduction and regularity.\hspace{-1cm}
\begin{defn}
  \label{def-red-all}
  Let $d$ be a \dterm, let $e$ be a subterm of $d$ and let $\alpha$ be an
  axiom assignment for~$d$.  For \dterms $e'$ the \dterm $\replace{d}{e}{e'}$
  is then obtained by \defname{\XC-reduction} from $d$ for $\alpha$ if $e
  \gtrc e'$, $\mgt_{\alpha}(e')$ is defined, and for all positions $p \in
  \pos(d)$ such that $d|_p = e$ it holds that $\ipt_{\alpha}(d,p) \subsumedBy
  \mgt_{\alpha}(e')$.  The \dterm~$d$ is called \defname{\XC-reducible}
  for~$\alpha$ if and only if there exists a \dterm~$e'$ such that
  $\replace{d}{e}{e'}$ is obtained by \XC-reduction from~$d$ for~$\alpha$.
  Otherwise, $d$ is called \defname{\XC-regular}.
\end{defn}

\vspace{-2pt}

\noindent
If $d'$ is obtained from~$d$ by \XC-reduction, then by
Theorem~\ref{thm-sr-ipt} and~\ref{thm-replace-csize-scsize} it follows that
$\mgt_{\alpha}(d) \subsumedBy \mgt_{\alpha}(d')$, $\csize(d) \geq \csize(d')$
and $\scsize(d) > \scsize(d')$.  \XC-regularity differs from well known
concepts of regularity in clausal tableaux (see, e.g.,
\cite{handbook:ar:haehnle}) in two respects: (1)~In the comparison of two
nodes on a branch (which is done by subsumption as in tableaux with universal
variables) for the upper node the stronger instantiated \IPT is taken and for
the lower node the more weakly instantiated \MGT. (2)~\XC-regularity is not
based on relating two nested subproofs, but on comparison of all occurrences
of a subproof with respect to all proofs that are smaller with respect to the
compaction ordering.

Proofs may involve applications of \Det where the conclusion $\P y$ is
actually independent from the minor premise $\P x$. Any axiom can then serve
as a trivial minor premise. Meredith expresses this with the symbol~$\n$ as
second argument of the respective \dterm.  Our function $\simpn$ simplifies
\dterms by replacing subterms with $\n$ accordingly on the basis of the
preservation of the \MGT.

\vspace{-3pt}

\begin{defn}
  \label{def-simp-n}
  If $d$ is a \dterm and $\alpha$ is an axiom assignment for~$d$, then the
  \defname{n-simplification of~$d$ with respect to~$\alpha$} is the \dterm
  $\simpn_{\alpha}(d)$, where $\simpn$ is the following function:
  $\simpn_{\alpha}(d) \eqdef d, \text{ if } d \text{ is a primitive \dterm}$;
  $\simpn_{\alpha}(\d(d_1,d_2)) \eqdef \d(\simpn_{\alpha'}(d_1),\n)$ \text{if
  } $\mgt_{\alpha'}\d(d_1,\n) = \mgt_{\alpha}\d(d_1,d_2)$, where $\alpha' =
          {\alpha \cup \{\n \mapsto \f{k}\}}$ for a fresh constant $\f{k}$;
          $\simpn_{\alpha}(\d(d_1,d_2)) \eqdef$\linebreak
          $\d(\simpn_{\alpha}(d_1),\simpn_{\alpha}(d_2))$,
            else.
\end{defn}

\vspace{-15pt}
\section{Properties of Meredith's Refined Proof}
\vspace{-7pt}
\enlargethispage{10pt}
\label{sec-properties}

\newcommand{\xatt}[1]{\textbf{#1}}

Our framework renders condensed detachment as a restricted form of the
CM. This view permits to consider the expanded proof structures as binary
trees or \dterms. On this basis we obtain a natural characterization of proof
properties in various categories, which seem to be the key towards reducing
the search space in ATP.  Table~\ref{tab-bigmer} shows such properties for
each of the 34 structurally different subproofs of proof~\PMER
(Fig.~\ref{fig-proof-mer}). Column~\xatt{M} gives the number of the subproof in
Fig.~\ref{fig-proof-mer}.  We use the following short identifiers for the
observed properties:

\newcommand{\tabatt}[1]{\scalebox{0.8}{\textbf{#1}}}
\newcommand{\colspc}{\hspace{1.8pt}}
\begin{table}[t]
  \centering
  \setlength{\tabcolsep}{0.93pt}
  \renewcommand{\arraystretch}{0.79}
    \scalebox{0.820}{\begin{tabular}{r@{\colspc}l@{\colspc}r@{\colspc}rrrrrccr@{\colspc}rrrrrccr@{\colspc}r@{\colspc}rrrr}
 & \tabatt{} & \tabatt{M} & \tabatt{DT} & \tabatt{DC} & \tabatt{DH} & \tabatt{DK$_{L}$} & \tabatt{DK$_{R}$} & \tabatt{DP} & \tabatt{DS} & \tabatt{DD} & \tabatt{DR} & \tabatt{TT} & \tabatt{TC} & \tabatt{TH} & \tabatt{TV} & \tabatt{TO} & \tabatt{RC} & \tabatt{MT} & \tabatt{MC} & \tabatt{IT$_{U}$} & \tabatt{IT$_{M}$} & \tabatt{IH$_{U}$} & \tabatt{IH$_{M}$}\\\midrule
1. & $1$ & \mer{1} & 0 & 0 & 0 & 0 & 0 & \textbullet & -- & 17 & 554 & 6 & 6 & 3 & 4 & \textbullet & \textbullet & 0 & 0 & 4451 & 203 & 18 & 11\\
2. & $\f{D}11$ &  & 1 & 1 & 1 & 1 & 1 & \textbullet & $1{=}1$ & 1 & 45 & 8 & 7 & 4 & 5 & \textbullet & \textbullet & 1 & 1 & 1640 & 220 & 17 & 12\\
3. & $\f{D}12$ &  & 2 & 2 & 2 & 1 & 2 & \textbullet & $1\lhd$ & 1 & 45 & 11 & 8 & 4 & 6 & \textbullet & \textbullet & 2 & 2 & 1881 & 252 & 17 & 12\\
4. & $\f{D}31$ &  & 3 & 3 & 3 & 2 & 2 & \textbullet & $\rhd1$ & 1 & 45 & 5 & 5 & 4 & 4 & \textcolor{black!30}{\textbullet} & \textbullet & 3 & 3 & 689 & 92 & 16 & 11\\
5. & $\f{D}4\mathrm{n}$ & \mer{2} & 4 & 4 & 4 & 3 & 2 & \textbullet & $\rhd\n$ & 1 & 45 & 4 & 4 & 3 & 3 & \textbullet & \textbullet & 4 & 4 & 688 & 91 & 15 & 10\\
6. & $\f{D}15$ &  & 5 & 5 & 5 & 3 & 2 & \textbullet & $1\lhd$ & 1 & 45 & 6 & 5 & 3 & 4 & \textbullet & \textbullet & 5 & 5 & 1667 & 198 & 15 & 10\\
7. & $\f{D}16$ &  & 6 & 6 & 6 & 3 & 3 & \textbullet & $1\lhd$ & 1 & 45 & 7 & 6 & 4 & 5 & \textbullet & \textbullet & 6 & 6 & 1802 & 208 & 16 & 11\\
8. & $\f{D}17$ &  & 7 & 7 & 7 & 3 & 4 & \textbullet & $1\lhd$ & 1 & 45 & 9 & 7 & 4 & 6 & \textbullet & \textbullet & 7 & 7 & 2648 & 303 & 16 & 11\\
9. & $\f{D}81$ &  & 8 & 8 & 8 & 3 & 4 & \textbullet & $\rhd1$ & 1 & 45 & 5 & 5 & 4 & 4 & \textcolor{black!30}{\textbullet} & \textbullet & 8 & 8 & 1032 & 119 & 15 & 10\\
10. & $\f{D}9\mathrm{n}$ & \mer{3} & 9 & 9 & 9 & 3 & 4 & \textbullet & $\rhd\n$ & 5 & 45 & 4 & 4 & 3 & 3 & \textbullet & \textbullet & 9 & 9 & 1031 & 118 & 14 & 9\\
11. & $\f{D}10.1$ & \mer{4} & 10 & 10 & 10 & 4 & 4 & \textbullet & $\rhd1$ & 2 & 37 & 4 & 4 & 3 & 3 & \textbullet & \textbullet & 10 & 10 & 448 & 60 & 13 & 9\\
12. & $\f{D}1.11$ &  & 11 & 11 & 11 & 4 & 4 & \textbullet & $1\lhd$ & 1 & 23 & 7 & 7 & 5 & 5 & \textbullet & \textbullet & 11 & 11 & 498 & 73 & 14 & 10\\
13. & $\f{D}1.12$ &  & 12 & 12 & 12 & 4 & 4 & \textbullet & $1\lhd$ & 1 & 23 & 12 & 8 & 5 & 6 & \textbullet & \textbullet & 12 & 12 & 1157 & 168 & 14 & 10\\
14. & $\f{D}1.13$ &  & 13 & 13 & 13 & 4 & 4 & \textbullet & $1\lhd$ & 1 & 23 & 10 & 9 & 6 & 7 & \textbullet & \textbullet & 13 & $\iv{12}{13}$ & 1050 & 159 & 15 & 11\\
15. & $\f{D}1.14$ &  & 14 & 14 & 14 & 4 & 5 & \textbullet & $1\lhd$ & 1 & 23 & 15 & 10 & 6 & 8 & \textbullet & \textbullet & 14 & $\iv{12}{14}$ & 1657 & 246 & 15 & 11\\
16. & $\f{D}15.1$ &  & 15 & 15 & 15 & 4 & 5 & \textbullet & $\rhd1$ & 1 & 23 & 9 & 8 & 5 & 6 & \textcolor{black!30}{\textbullet} & \textbullet & 15 & $\iv{12}{15}$ & 684 & 100 & 14 & 10\\
17. & $\f{D}16.\mathrm{n}$ & \mer{5} & 16 & 16 & 16 & 4 & 5 & \textbullet & $\rhd\n$ & 2 & 23 & 8 & 7 & 4 & 5 & \textbullet & \textbullet & 16 & $\iv{12}{16}$ & 683 & 99 & 13 & 9\\
18. & $\f{D}17.1$ & \mer{6} & 17 & 17 & 17 & 4 & 5 & \textbullet & $\rhd1$ & 3 & 18 & 7 & 6 & 3 & 4 & \textbullet & \textbullet & 17 & $\iv{12}{17}$ & 395 & 56 & 12 & 8\\
19. & $\f{D}18.11$ & \mer{7} & 28 & 18 & 18 & 5 & 5 & -- & $\rhd$ & 1 & 14 & 7 & 6 & 4 & 4 & \textbullet & \textbullet & 14 & $\iv{12}{14}$ & 209 & 61 & 11 & 9\\
20. & $\f{D}19.1$ & \mer{8} & 29 & 19 & 19 & 6 & 5 & -- & $\rhd1$ & 2 & 14 & 9 & 8 & 5 & 5 & \textbullet & \textbullet & 15 & $\iv{12}{15}$ & 132 & 38 & 10 & 8\\
21. & $\f{D}1.20$ & \mer{10} & 30 & 20 & 20 & 6 & 5 & -- & $1\lhd$ & 2 & 10 & 12 & 9 & 5 & 6 & \textbullet & \textbullet & 16 & $\iv{12}{16}$ & 158 & 47 & 10 & 8\\
22. & $\f{D}21.21$ &  & 61 & 21 & 21 & 6 & 5 & -- & $=$ & 1 & 5 & 10 & 9 & 5 & 6 & \textcolor{black!30}{\textbullet} & \textbullet & $\iv{23}{33}$ & $\iv{12}{17}$ & 53 & 16 & 9 & 7\\
23. & $\f{D}22.\mathrm{n}$ & \mer{11} & 62 & 22 & 22 & 6 & 5 & -- & $\rhd\n$ & 1 & 5 & 9 & 8 & 4 & 5 & \textbullet & \textbullet & $\iv{23}{34}$ & $\iv{12}{18}$ & 52 & 15 & 8 & 6\\
24. & $\f{D}17.23$ & \mer{12} & 79 & 23 & 23 & 6 & 5 & -- & $\lhd$ & 2 & 5 & 9 & 8 & 4 & 5 & \textbullet & \textbullet & $\iv{23}{51}$ & $\iv{12}{23}$ & 57 & 16 & 7 & 5\\
25. & $\f{D}24.18$ & \mer{13} & 97 & 24 & 24 & 6 & 5 & -- & $\rhd$ & 2 & 2 & 7 & 6 & 4 & 4 & \textbullet & \textbullet & $\iv{23}{69}$ & $\iv{12}{24}$ & 27 & 17 & 6 & 5\\
26. & $\f{D}20.10$ & \mer{9} & 39 & 20 & 20 & 7 & 5 & -- & $\rhd$ & 2 & 4 & 3 & 2 & 2 & 2 & \textbullet & -- & 8 & 6 & 27 & 7 & 6 & 4\\
27. & $\f{D}24.26$ & \mer{14} & 119 & 25 & 24 & 7 & 5 & -- & $>_{\mathrm{c}}$ & 2 & 3 & 5 & 5 & 3 & 3 & \textbullet & \textbullet & $\iv{23}{91}$ & $\iv{12}{25}$ & 24 & 7 & 6 & 4\\
28. & $\f{D}10.27$ & \mer{15} & 129 & 26 & 25 & 7 & 5 & -- & $\lhd$ & 1 & 2 & 3 & 3 & 3 & 2 & \textbullet & \textbullet & $\iv{23}{101}$ & $\iv{12}{26}$ & 19 & 12 & 6 & 5\\
29. & $\f{D}18.28$ & \mer{16} & 147 & 27 & 26 & 7 & 5 & -- & $\lhd$ & 2 & 2 & 5 & 5 & 4 & 3 & \textbullet & \textbullet & $\iv{23}{36}$ & $\iv{12}{26}$ & 19 & 12 & 6 & 5\\
30. & $\f{D}29.29$ &  & 295 & 28 & 27 & 7 & 6 & -- & $=$ & 1 & 1 & 10 & 7 & 5 & 4 & \textbullet & \textbullet & $\iv{23}{239}$ & $\iv{12}{27}$ & 13 & 13 & 5 & 5\\
31. & $\f{D}25.30$ &  & 393 & 30 & 28 & 7 & 7 & -- & $<_{\mathrm{c}}$ & 1 & 1 & 7 & 7 & 5 & 4 & \textbullet & \textbullet & $\iv{23}{121}$ & $\iv{12}{29}$ & 13 & 13 & 5 & 5\\
32. & $\f{D}31.25$ & \mer{17} & 491 & 31 & 29 & 7 & 7 & -- & $\rhd$ & 0 & 1 & 5 & 5 & 3 & 3 & \textbullet & \textbullet & $\iv{23}{191}$ & $\iv{12}{30}$ & 5 & 5 & 3 & 3\\
33. & $\f{D}27.26$ & \mer{18} & 159 & 26 & 25 & 7 & 5 & -- & $\rhd$ & 0 & 1 & 3 & 3 & 3 & 2 & \textbullet & \textbullet & 15 & 11 & 3 & 3 & 3 & 3\\
34. & $\f{D}10.10$ & \mer{19} & 19 & 10 & 10 & 4 & 4 & -- & $=$ & 0 & 1 & 2 & 2 & 2 & 2 & \textbullet & \textbullet & 7 & 6 & 2 & 2 & 2 & 2\\
\end{tabular}
}
  \vspace{-2pt}
  \caption{Properties of all subproofs of the proof \PMER
    \cite{meredith:notes:1963} shown in Fig.~\ref{fig-proof-mer}.}
  \label{tab-bigmer}
  \vspace{-30pt}
\end{table}

\boldpar{Structural Properties of the \DTerm.}
\label{sec-prop-struct}
These properties refer to the respective subproof as \dterm or full binary
tree.
\xatt{DT}, \xatt{DC}, \xatt{DH}: Tree size, compacted size, height.
\xatt{DK$_L$}, \xatt{DK$_R$}: ``Successive height'', that is, the maximal
number of successive edges going to the left (right, resp.) on any path from
the root to a leaf. \xatt{DP}: Is ``prime'', that is, \xatt{DT} and \xatt{DC}
are equal.  \xatt{DS}: Relationship between the subproofs of major and minor
premise. Identity is expressed with $=$, the subterm and superterm
relationships with $\lhd$ and $\rhd$, resp., and the compaction ordering
relationship (if none of the other relationships holds) with
$\mathrel{<_{\mathrm{c}}}$ and $\mathrel{>_{\mathrm{c}}}$. In addition it is
indicated if a subproof is an axiom or $\n$. \xatt{DD}: ``Direct sharings'',
that is, the number of incoming edges in the DAG representation of the overall
proof of all theorems.  \xatt{DR}: ``Repeats'', that is, the total number of
occurrences in the set of expanded trees of all roots of the DAG.

\boldpar{Properties of the \MGT.}  These properties refer to the argument term
of the \MGT of the respective subproof.  \xatt{TT}, \xatt{TH}: Tree size
(defined as for \dterms) and height.  \xatt{TV}: Number of different variables
occurring in the term.  \xatt{TO}: Is ``organic''
\cite{luk:tarski:aussagenkalkuel:1930}, that is, the argument term has no
strict subterm~$s$ such that $\P(s)$ itself is a theorem.  We call an atom
\emph{weakly organic} (indicated by a gray bullet) if it is not organic and
the argument term is of the form $\i(p,t)$ where $p$ is a variable that does
not occur in the term $t$ and $\P(t)$ is organic. For axiomatizations of
fragments of propositional logic, \name{organic} can be checked
by a SAT solver.

\boldpar{Regularity.} \xatt{RC}: The respective subproof as \dterm is
\XC-regular (see Def.~\ref{def-red-all}).

\boldpar{Comparisons with all Proofs of the \MGT.}  These properties relate to
the set of all proofs (as \dterms) of the \MGT of the respective subproof.
\xatt{MT}, \xatt{MC}: Minimal tree size and minimal compacted size of a proof.
These values can be hard to determine such that in Table~\ref{tab-bigmer} they
are often only narrowed down by an integer interval.  To determine them, we
used the proof \PMER, proofs obtained with techniques described in
Sect.~\ref{sec-experiments}, and enumerations of all \dterms with defined \MGT
up to a given tree size or compacted size.

\enlargethispage{6pt}

\boldpar{Properties of Occurrences of the \IPTs.}  The respective subproof has
\xatt{DR} occurrences in the set of expanded trees of the roots of the DAG,
where each occurrence has an \IPT.  The following properties refer to the
multiset of argument terms of the \IPTs of these occurrences.
\xatt{IT$_U$}, \xatt{IT$_M$}: Maximal tree size and rounded median of the tree
size.  \xatt{IH$_U$}, \xatt{IH$_M$}: Maximal height and rounded median of the
height.  In Table~\ref{tab-bigmer} these values are much larger than those of
the corresponding columns for the \MGT, i.e, \xatt{TT} and \xatt{TH},
illustrating Prop.~\ref{prop-ipt-subsumedby-mgt}.

\renewcommand{\mer}[1]{\textit{M#1}}

\section{First Experiments}
\label{sec-experiments}
\vspace{-11pt}
\enlargethispage{30pt}

First experiments based on the framework developed in the previous sections
are centered around the generation of lemmas where not just formulas but, in
the form of \dterms, also proofs are taken into account.  This leads in
general to preference of small proofs and to narrowing down the search space
by restricted structuring principles to build proofs.
The experiments indicate novel calculi -- for now at an early stage of
development -- which combine aspects from lemma-based generative, or
bottom-up, methods such as hyperresolution and hypertableaux with
structure-based approaches that are typically used in an analytic, or
goal-directed, way such as the connection method.  In addition, ways of using
lemma generation as preprocessing for theorem proving, in particular to obtain
short proofs, are suggested.
These techniques resulted in a further refinement of \Lukasiewicz's proof
\cite{luk:1948} of the completeness of his single axiom for the implicational
fragment, whose compacted size is by one smaller than that of Meredith's
refinement \cite{meredith:notes:1963} and by two than \Lukasiewicz's original
proof.

\newcommand{\PRIMES}{\textit{PrimeCore}\xspace}
\newcommand{\PSUB}{\textit{ProofSubproof}\xspace}
\newcommand{\FPRIMES}{\textit{PrimeCore}}
\newcommand{\FPSUB}{\textit{ProofSubproof}}
\begin{table}[t]
  \centering
  \scriptsize
   {\renewcommand{\arraystretch}{0.83}
\begin{tabular}{r@{\hspace{6pt}}lrr@{\hspace{0.5em}}|@{\hspace{0.5em}}%
     lr@{\hspace{0.5em}}|@{\hspace{0.5em}}r@{\hspace{6pt}}rr}
  & \textit{Lemmas} & \textit{\#} & \textit{Time} &
  \textit{Prover} & \textit{Time} & 
  \xatt{DC} & \xatt{DT} &  \xatt{DH}\\\midrule
  1. &&&& \Lukasiewicz$\!\!^*$ & & 32 & 435 & 29\\
  2. &&&& Meredith && 31 & 491 & 29\\
  3.  &&&& \ProverN & 37~s & 94 & 304,890 & 40\\
  4.  &&&& \ProverN$\!\!^*$ & 37~s & 83 & 8,217 & 38\\
  5.  &&&& \ProverN$\!\!^*$ depth $\leq 7$ & 6~s & 102 & 19,113 & 48\\
  6. & \FPRIMES(17) & 17 &  & \ProverN$\!\!^*$ & 30~s & 44 & 763 & 28\\  
  7. & \FPSUB(93,7) & 291 & 78 s & \ProverN$\!\!^*$ & 3~s & 51 & 1,405 & 31\\
  8. & \FPSUB(93,7) & 291 & 78 s & \CMProver & 2~s & 30 & 394 & 29\\
  9. & \FPSUB(100,8) & 330 & 94~s & \CMProver & 4~s & 30 & 535 & 29\\
  10. &&&& Reduction of (8.) & & 48 & 191 & 24\\
\end{tabular}}
\caption{Proof dimensions of various proofs of problem~\PSYLL.}
\label{tab-exp}
\vspace{-24pt}
\end{table}

Table~\ref{tab-exp} shows compacted size \xatt{DC}, tree size \xatt{DT} and
height \xatt{DH} of various proofs of~\PSYLL.  Asterisks indicate that
n-simplification was applied with reducing effect on the system's proof.
Proof~(1.) is the one by \Lukasiewicz \cite{luk:1948}, translated into
condensed detachment, proof~(2.) is proof~\PMER (Fig.~\ref{fig-proof-mer})
\cite{meredith:notes:1963}. Rows~(3.)--(5.) show results from \ProverN, where
in~(5.)  the value of \texttt{max\_depth} was limited to~7, motivated by
column~\xatt{TH} of Table~\ref{tab-bigmer}.  Proof~(4.)  illustrates the
effect of n-simplification.\footnote{All machine results refer to a system
  with Intel i7-8550U CPU and 16~GB RAM.
  Results for further systems: \KRHyper$\!\!^*$
  \cite{cw-ekrhyper}: 1.610~s, \xatt{DC}:~73; \name{E~2.5} \cite{eprover}:
  30~s, proof length 91; \name{Vampire~5.4.1} \cite{vampire} \texttt{--mode
    casc -t 300}: 128~s, proof  length~144.}
For proofs~(6.)--(9.) additional axioms were supplied to \ProverN and
\mbox{\CMProver} \cite{cw-mathlib,cw-pie:2016,cw-pie:2020}, a goal-directed
system that can be described by the CM.   Columns indicate the lemma
computation method, the number of lemmas supplied to the prover and
the time used for lemma computation.
Method \PRIMES adds the MGTs of subproof~18 from Table~\ref{tab-bigmer} and
all its subproofs as lemmas. Subproof~18 is the largest subproof of proof
\PMER that is prime and can be characterized on the basis of the axiom --
almost uniquely -- as a proof that is prime, whose MGT has no smaller prime
proof and has the same number of different variables as the axiom, i.e., 4,
and whose size, given as parameter, is~17.
Method \PSUB is based on detachment steps with a \dterm and a subterm of
it as proofs of the premises, which, as column~\xatt{DS} of
Table~\ref{tab-bigmer} shows, suffices to justify all except of two proof
steps in \PMER.  It proceeds in some analogy to the given clause algorithm on
lists of \dterms: If~$d$ is the given \dterm, then the \emph{inferred \dterms}
are all \dterms that have a defined \MGT and are of the form $\d(d,e)$ or
$\d(e,d)$, where $e$ is a subterm of $d$. To determine which of the inferred
\dterms are kept, values from Table~\ref{tab-bigmer} were taken as guide,
including \xatt{RC} and \xatt{TO}.  The first parameter of \PSUB is the number
of iterations of the ``given \dterm loop''.
Proof~(9.)  can be combined with \Peirce and \Syll to the overall proof with
compacted size~32, one less than \PMER. The maximal value of \xatt{DK$_L$} is
shown as second parameter, because, when limited to~$7$, proof~(9.) cannot be
found.  Proof~(10.), which has a small tree size, was obtained from~(8.) by
rewriting subproofs with a variation of \XC-reduction that rewrites single
term occurrences, considering also \dterms from a precomputed table of small
proofs.

\vspace{-11pt}
\section{Conclusion}
\label{sec-conclusion}
\vspace{-7pt}
\enlargethispage{9pt}

Starting out from investigating \Lukasiewicz's classic formal proof
\cite{luk:1948}, via its refinement by Meredith \cite{meredith:notes:1963} we
arrived at a formal reconstruction of Meredith's condensed detachment as a
special case of the CM. The resulting formalism yields proofs as objects of a
very simple and common structure: full binary trees which, in the tradition of
term rewriting, appear as terms, \Dterms, as we call them.  To form a full
proof, formulas are associated with the nodes of \Dterms: axioms with the
leaves and lemmas with the remaining nodes, implicitly determined from the
axioms through the node position and unification. The root lemma is the most
general proven theorem.  Lemmas also relate to compressed representations of
the binary trees, for example as DAGs, where the re-use of a lemma directly
corresponds to sharing the structure of its subproof.
For future work we intend to position our approach also in the context of
earlier works on proofs, proof compression and lemma introduction, e.g.,
\cite{woltzenlogel:cutintro:2010,vienna:qcuts:2014}, and think of compressing
\DTerms in forms that are stronger than DAGs, e.g., by tree grammars
\cite{lohrey:survey:2015}.

The combination of formulas and explicitly available proof structures
naturally leads to theorem proving methods that take structural aspects into
account, in various ways, as demonstrated by our first experiments.  This goes
beyond the common clausal tableau realizations of the CM, which in essence
operate by enumerating uncompressed proof structures.
The discussed notions of regularity and lemma generation methods seem
immediately suited for further investigations in the context of first-order
theorem proving in general. For other aspects of the work we plan a stepwise
generalization by considering further single axioms for the implicational
fragment \textbf{IF}
\cite{luk:tarski:aussagenkalkuel:1930,luk:1948,ulrich:single:2016}, single
axioms and axiom pairs for further logics \cite{ulrich:single:2016}, the about
200~condensed detachment problems in the LCL domain of the TPTP, problems
which involve multiple non-unit clauses, and adapting \dterms to a variation
of binary resolution instead of detachment.
In the longer run, our approach
aims at providing a basis for approaches to theorem proving with machine
learning (e.g. \cite{FaeKU:2020,enigma:2020}). With the reification of proof
structures more information is available as starting point.  As indicated with
our exemplary feature table for Meredith's proof, structural properties are
considered thereby from a global point of view, as a source for narrowing down
the search space in many different ways in contrast to just the common local
view ``from within a structure'', where the narrowing down is achieved for
example by focusing on a ``current branch'' during the construction of a
tableau.  A general lead question opened up by our setting is that for
exploring relationships between properties of proof structures and the
associated formulas in proofs of meaningful theorems.  One may expect that
characterizations of these relationships can substantially restrict the search
space for finding proofs.

\clearpage
\bibliography{biblukas01}

\bibliographystyle{splncs04}

\clearpage
\appendix

\setlength{\abovedisplayskip}{10.0pt plus 2.0pt minus 5.0pt}%
\setlength{\belowdisplayskip}{10.0pt plus 2.0pt minus 5.0pt}%
\setlength{\abovedisplayshortskip}{0.0pt plus 3.0pt}
\setlength{\belowdisplayshortskip}{6.0pt plus 3.0pt minus 3.0pt}%

\section{Proofs of Claims in the Paper, Additional Examples, and
  Refined Formal Proofs of the Completeness of \Lukasiewicz's Single Axiom}
\label{sec-appendix}

\subsection{Supplementary Material for Section~\ref{sec-subst-base}}

The following example shows for a given \Dterm the set of associated pairings
(Def.~\ref{def-pairing}) with its most general unifier (Defs.~\ref{def-mgu}
and~\ref{def-mgu-content}), as well as the \IPT and \MGT for a specific
position in the \Dterm (Def.~\ref{def-ipt-mgt}).
\begin{exampapp}
  \label{examp-ipt-mgt}
  Let $\alpha$ be an axiom assignment that maps the primitive \dterm~$1$ to
  the canonical representation of the axiom \Simp.  That is, \[\alpha \eeqdef
  \{1 \mapsto \c(\vx{1}{\emptypos}, \c(\vx{2}{\emptypos},
  \vx{1}{\emptypos}))\}.\]  Consider the \dterm $d \eeqdef \D(\D(1,1),1)$.
  Then $\pos(d) = \{\emptypos, 1, 11, 12, 2\}$ and
  \[\begin{array}{l@{\hspace{0.5em}}c@{\hspace{0.5em}}l}
  \pairing_{\daxioms}(d,\emptypos) & = & \{\vy_{1},\, \c(\vy_{2}, \vy_{\emptypos})\}.\\
  \pairing_{\daxioms}(d,1) & = & \{\vy_{1.1},\, \c(\vy_{1.2}, \vy_1)\}.\\
  \pairing_{\daxioms}(d,1.1) & = & \{\vy_{1.1},\, \c(\vx{1}{1.1}, \c(\vx{2}{1.1}, \vx{1}{1.1}))\}.\\
  \pairing_{\daxioms}(d,1.2) & = & \{\vy_{1.2},\, \c(\vx{1}{1.2}, \c(\vx{2}{1.2}, \vx{1}{1.2}))\}.\\
  \pairing_{\daxioms}(d,2) & = & \{\vy_{2},\, \c(\vx{1}{2}, \c(\vx{2}{2}, \vx{1}{2}))\}.\\
  \end{array}
  \]
  Let $\sigma \eeqdef \mgu(\{\pairing_{\daxioms}(d, q) \mid q \in
  \pos(d)\}))$. We can then calculate that
  \[\begin{array}{l@{\hspace{0.5em}}c@{\hspace{0.5em}}ll}
  \sigma & \variant & \{
  & \vy_{\emptypos} \mapsto \c(\vx{1}{1.2}, \c(\vx{2}{1.2}, \vx{1}{1.2})),\\
  &&& \vy_{1} \mapsto \c(\c(\vx{1}{2}, \c(\vx{2}{2}, \vx{1}{2})), \c(\vx{1}{1.2}, \c(\vx{2}{1.2}, \vx{1}{1.2}))),\\
  &&& \vy_{1.1} \mapsto \c(\c(\vx{1}{1.2}, \c(\vx{2}{1.2}, \vx{1}{1.2})), \c(\c(\vx{1}{2}, \c(\vx{2}{2}, \vx{1}{2})), \c(\vx{1}{1.2}, \c(\vx{2}{1.2}, \vx{1}{1.2})))),\\
  &&& \vy_{1.2} \mapsto \c(\vx{1}{1.2}, \c(\vx{2}{1.2}, \vx{1}{1.2})),\\
  &&& \vy_{2} \mapsto \c(\vx{1}{2}, \c(\vx{2}{2}, \vx{1}{2})),\\
  &&& \vx{1}{1.1} \mapsto \c(\vx{1}{1.2}, \c(\vx{2}{1.2}, \vx{1}{1.2})),\\
  &&& \vx{2}{1.1} \mapsto \c(\vx{1}{2}, \c(\vx{2}{2}, \vx{1}{2}))\; \}.
  \end{array}
  \]
  Let $d' \eeqdef d|_1$, that is, $d'$ is the subterm of $d$ at
  position~$1$. Then $d' = \D(1,1)$, $\pos(d') = \{\emptypos, 1, 2\}$, and
  \[\begin{array}{l@{\hspace{0.5em}}c@{\hspace{0.5em}}l}
  \pairing_{\daxioms}(d',\emptypos) & = & \{\vy_{1},\, \c(\vy_{2}, \vy_{\emptypos})\}.\\
  \pairing_{\daxioms}(d',1) & = & \{\vy_{1},\, \c(\vx{1}{1}, \c(\vx{2}{1}, \vx{1}{1}))\}.\\
  \pairing_{\daxioms}(d',2) & = & \{\vy_{2},\, \c(\vx{1}{2}, \c(\vx{2}{2}, \vx{1}{2}))\}.\\
  \end{array}
  \]
  Let $\sigma' \eeqdef \mgu(\{\pairing_{\daxioms}(d', q) \mid q \in
  \pos(d)\}))$. Then
  \[\begin{array}{l@{\hspace{0.5em}}c@{\hspace{0.5em}}ll}
  \sigma' & \variant & \{
  & \vy_{\emptypos} \mapsto \c(\vx{2}{1}, \c(\vx{1}{2}, \c(\vx{2}{2}, \vx{1}{2}))),\\
  &&& \vy_{1} \mapsto \c(\c(\vx{1}{2}, \c(\vx{2}{2}, \vx{1}{2})), \c(\vx{2}{1}, \c(\vx{1}{2}, \c(\vx{2}{2}, \vx{1}{2})))),\\
  &&& \vy_{2} \mapsto \c(\vx{1}{2}, \c(\vx{2}{2}, \vx{1}{2})),\\
  &&& \vx{1}{1} \mapsto \c(\vx{1}{2}, \c(\vx{2}{2}, \vx{1}{2}))\; \}.
  \end{array}
  \]  
  Now $\ipt(d,1)$ and $\mgt(d|_1)$ can be determined as follows, where we
  supplement the values obtained by applying the displayed unifiers with
  variants that have variable names $p,q,r,s$ and are easier to read:
  \[
  \begin{array}{l@{\hspace{0.5em}}c@{\hspace{0.5em}}l}
    \ipt(d,1) & = & \P(y_1\sigma)\\
    & \variant & \P(\c(\c(\vx{1}{2}, \c(\vx{2}{2}, \vx{1}{2})), \c(\vx{1}{1.2}, \c(\vx{2}{1.2},
    \vx{1}{1.2}))))\\
    & \variant & \P(\c(\c(p, \c qp), \c(r, \c sr))).\\[1ex]
    \mgt(d|_1) & = & \mgt(d')\\
    & = & \ipt(d',\emptypos)\\
    & = & \P(y_{\emptypos}\sigma')\\
    & \variant & \P(\c(\vx{2}{1}, \c(\vx{1}{2}, \c(\vx{2}{2}, \vx{1}{2}))))\\
    & \variant & \P(\c(p, \c(q, \c rq))).
  \end{array}
  \]
  That $\ipt(d,1) \subsumedBy \mgt(d|_1)$ as claimed by
  Prop.~\ref{prop-ipt-subsumedby-mgt} holds follows since
  \[\c(\c(p, \c qp), \c(r, \c sr)) \strictlySubsumedBy \c(p, \c(q, \c rq)).\]
  Side remark: In this simple example it holds that $\mgt(d) \variant \P(\c(p,
  \c(q, p)))$, that is, the \MGT of $d$ is a variant of the
  axiom. 
\end{exampapp}

\medskip
\noindent
The following example illustrates the application of $\f{shift}$
(Def.~\ref{def-sshift}).
\begin{exampapp}
  \[\begin{array}{l@{\hspace{0.5em}}c@{\hspace{0.5em}}l}
  \c(\vx{1}{\emptypos},\vx{2}{\emptypos})\sshift{1.1.2.1} & = &
  \c(\vx{1}{1.1.2.1},\vx{2}{1.1.2.1}).\\[3pt]
  \c(\vy_{2.1},\vy_{2.1.2}))\sshift{1.1} & = &
  \c(\vy_{1.1.2.1},\vy_{1.1.2.1.2}).
  \end{array}\]
  In the second example, observe that position~$2.1.2$ refers to the right
  child of position~$2.1$. After applying $\sshift{1.1}$, it is
  position~$1.1.2.1.2$ that, again, refers to the right child of
  position~$1.1.2.1$.
\end{exampapp}
Applying a $\sshift{p}$ substitution to a term always yields a variant, as
stated in the following proposition.
\begin{propapp}
  \label{prop-shift-variant}
  For all terms~$s$ such that $\vars(s) \subseteq \posvar$ and positions~$p$
  it holds that
  \[s\; \variant\; s\sshift{p}.\]
\end{propapp}
\begin{proof}
  Easy to see. \qed
\end{proof}
The following proposition shows an interplay of $\pairing$ and $\f{shift}$
that is used later in proofs.
\begin{propapp}
  \label{prop-tree-renaming-add}
  Let $d$ be a \dterm, let $p$ be a position in $\pos(d)$ and let~$\daxioms$
  be an axiom assignment for $d$. Then
  \[\begin{array}{r@{\hspace{0.5em}}l}
    & y_{\emptypos}\mgu(\{\pairing_{\daxioms}(d|_p, q) \mid q \in \pos(d|_p)\})
    \sshift{p}\\
    = & y_{p}\mgu(\{\pairing_{\daxioms}(d, q) \mid q \in \pos(d) \text{
      and } p \leq q\}).
  \end{array}
  \]
\end{propapp}
\begin{proof}
  Easy to see. \qed
\end{proof}
Sometimes it is useful to refer to all variables associated with positions or
associated with members of a given set of positions, regardless of whether
they are of the form $\vy_p$ or $\vx{i}{p}$. The following definition provides
a notation for this.
\begin{defnapp}
  \ \par
  \subdefnapp{def-vpos}
  $\posvar\; \eqdef\; \{\vy_p \mid p \text{ is a position }\} \cup
  \{\vx{i}{p} \mid p \text{ is a position and } i \geq 1\}$.
  
  \subdefnapp{def-vpos-p} For all sets~$P$ of positions define
  \[\posvar(P)\; \eqdef\; \{\vy_p \mid p \in P\}
  \cup \{\vx{i}{p} \mid p \in P \text{ and } i \geq 1\}.\]
\end{defnapp}

\medskip
\noindent
We are now ready to prove Prop.~\ref{prop-ipt-subsumedby-mgt}.

\begin{propref}{}
  {\ref{prop-ipt-subsumedby-mgt}}
  For all \dterms~$d$, positions $p \in \pos(d)$ and axiom assignments
  $\daxioms$ for $d$ it holds that \[\ipt_{\daxioms}(d,p) \subsumedBy
  \mgt_{\daxioms}(d|_p).\]
\end{propref}
\begin{proof}
  \prlReset{prop-ipt-subsumedby-mgt} Can be shown in the following steps,
  explained below:
  \[
  \begin{array}{r@{\hspace{0.5em}}c@{\hspace{0.5em}}l@{\hspace{0.5em}}l}
  \prl{1} & & \ipt_{\daxioms}(d,p)\\
  \prl{2} & = & \P(y_{p}\mgu(\{\pairing_{\daxioms}(d, q) \mid q \in \pos(d)\}))\\
  \prl{3} & \subsumedBy & \P(y_{p}\mgu(\{\pairing_{\daxioms}(d,q) \mid q \in \pos(d) \text{ and } p \leq q\}))\\
  \prl{4} & = & \P(y_{\emptypos}\mgu(\{\pairing_{\daxioms}(d|_p, q) \mid q \in \pos(d|_p)\})\sshift{p})\\
  \prl{5} & \variant & \P(y_{\emptypos}\mgu(\{\pairing_{\daxioms}(d|_p, q) \mid q \in \pos(d|_p)\}))\\
  \prl{6} & = & \ipt_{\daxioms}(d|_p, \emptypos)\\
  \prl{7} & = & \mgt_{\daxioms}(d|_p).\\
  \end{array}
  \]
   Step~\pref{3} follows easily from the definition of most general unifier.
   Step~\pref{4} is justified by Prop.~\ref{prop-tree-renaming-add},
   step~\pref{5} by Prop.~\ref{prop-shift-variant}.  The remaining steps are
   obtained by expanding and contracting definitions.
   \qed
\end{proof}
By universally closing the atoms on both sides of
Prop.~\ref{prop-ipt-subsumedby-mgt} we can relate MGT and IPT
by entailment.
\begin{propapp}
  \label{prop-mgt-entails-ipt} For all \dterms~$d$ and positions $p \in
  \pos(d)$ it holds that
  $\UCN{\mgt_{\daxioms}(d|_p)}\; \entails\; \UCN{\ipt_{\daxioms}(d,p)}.$
\end{propapp}
\begin{proof} Follows from
  Prop.~\ref{prop-ipt-subsumedby-mgt}.  \qed
\end{proof}

Lemma~\ref{lem-core} and Theorem~\ref{thm-sem-mgt} can be proven as follows.

\begin{lemref}{}
  {\ref{lem-core}}
  Let $d$ be a \dterm and let $\daxioms$ be an axiom assignment for $d$. Then
  for all $p \in \pos(d)$ it holds that:

  \smallskip
  \subpropref{lem-core-leaf} If $p \in \leafpos(d)$, then
  \[\UCM{\P(\daxioms(d|_p))}\; \entails\; \ipt_{\daxioms}(d,p).\]

  \subpropref{lem-core-inner} If $p \in \innerpos(d)$, then
  \[\Det \land \ipt_{\daxioms}(d,p.1) \land \ipt_{\daxioms}(d,p.2)\; \entails\; \ipt_{\daxioms}(d,p).\]
\end{lemref}

\begin{proof}
  Let $\sigma = \mgu(\{\pairing_{\daxioms}(d, q) \mid q \in \pos(d)\})$ and
  assume it is defined.
  
  \smallskip
  (\ref{lem-core-leaf}) From Def.~\ref{def-ipt} and Def.~\ref{def-pairing} we
  can conclude $\ipt_{\daxioms}(d,p)\, =\, \P(\vy_{p}\sigma)\, =$\linebreak
  $\P(\daxioms(d|_p)\sshift{p}\sigma)\, \subsumedBy\, \P(\daxioms(d|_p))$,
  which implies the proposition to be proven.

  \smallskip
  (\ref{lem-core-inner}) From Def.~\ref{def-ipt} and Def.~\ref{def-pairing} we
  can conclude $\ipt(d,p.1)\, =\, \P(\vy_{p.1}\sigma)\, =\, \P(\c(\vy_{p.2},
  \vy_p)\sigma)$, $\ipt(d,p.2)\, =\, \P(\vy_{p.2}\sigma)$, and $\ipt(d,p)\,
  =\, \P(\vy_{p}\sigma)$. Hence, we can rephrase the proposition statement as
  \[\Det \land \P(\c(\vy_{p.2},
  \vy_p)\sigma) \land \P(\vy_{p.2}\sigma)\; \entails\; \P(\vy_{p}\sigma).\] By
  expanding the definition of~$\Det$ and rearranging formula components, this
  entailment can be brought into the following form that obviously holds, as
  its right side is obtained from instantiating universal quantifiers on the
  left side:
  \[\forall xy\, (\P x \land \P\c xy \imp \P y)\;
  \entails\; \P(\vy_{p.2}\sigma) \land \P(\c(\vy_{p.2}, \vy_p)\sigma) \imp
  \P(\vy_{p}\sigma).\vspace{-2ex}\]
  \qed
\end{proof}

\begin{thmref}{}
  {\ref{thm-sem-mgt}}
  Let $d$ be a \dterm and let $\daxioms$ be an axiom assignment for $d$.  Then
  \[\Det \land \bigwedge_{p \in \leafpos(d)}\hspace{-1.2em} \UCM{\P(\daxioms(d|_p))}
  \;\entails\; \UCN{\mgt_{\daxioms}(d)}.\]
\end{thmref}
\begin{proof}
 By induction on the structure of $d$ it follows from Lemma~\ref{lem-core}
 that \[\Det \land \bigwedge_{p \in \leafpos(d)}\hspace{-1.2em}
 \UCM{\P(\daxioms(d|_p))}
 \;\entails\; \ipt_{\daxioms}(d,\emptypos).\] Contracting the definition of
 $\mgt$, the right side of this entailment can be written as
 $\mgt_{\daxioms}(d)$. Since the left side of the entailment has no free
 variables, we can replace the right side with its universal closure and
 obtain the statement to be proven. \qed
\end{proof}

\subsection{Supplementary Material for Section~\ref{sec-red}}

The relation $d \gtrc e$ (Def.~\ref{def-c-orderings}) can equivalently be
characterized as $\{f \in \DC \mid d \supterm f\} \supset \{f \in \DC \mid \{f
\in \DC \mid e \supterm f\}$.  Hence, the underlying comparison is for $\geqc$
with respect to the non-strict superset relationship and for $\gtrc$ the
strict superset relationship.  The $\geqc$ relation is a preorder on the set
of \dterms, while $\gtrc$ is a strict partial order.  The subterm relationship
includes the compaction orderings, as noted by the following proposition.

\begin{propapp}
  \label{prop-gecq-supterm-all}
  For all \dterms $d,e,f$ it holds that
  \smallskip
  
  \subpropapp{prop-geqc-supterm} If $d \suptermq e $, then $d \geqc e$.

  \subpropapp{prop-gtrc-supterm} If $d \supterm e $ and $d$ is not of the form
  $\D(l_1,l_2)$ where both of $l_1, l_2$ are primitive \dterms, then
  $d \gtrc e$.
  
  \subpropapp{prop-geqc-supterm-trans} If $d \suptermq e $ and $e \geqc f$, then
  $d \geqc f$.

  \subpropapp{prop-gtcr-supterm-trans} If $d \suptermq e $ and $e \gtrc f$, then
  $d \gtrc f$.

\end{propapp}

\begin{proof}
  Easy to verify. \qed
\end{proof}
According to Propositions~\ref{prop-geqc-supterm} and~\ref{prop-gtrc-supterm}
the subterm relationship includes the compaction orderings, with an exception,
as stated in the precondition of Prop.~\ref{prop-gtrc-supterm}. (An example
for the exception is $\D(1,1) \supterm 1$ but $\D(1,1) \not \gtrc 1$.)
However, $d \geqc e$ or $d \gtrc e$ also holds in cases where $d \not
\suptermq e$, as shown in the following example.
\begin{exampapp}
  \label{examp-co}
  The following table shows \dterms~$d$ and $e$ where $d \geqc e$ or $d \gtrc
  e$ holds but $d \not \suptermq e$.  \prlReset{examp-co} The respective
  values of $\{f \in \DC \mid d \supterm f\}$ and $\{f \in \DC \mid e \supterm
  f\}$ according to the definition of $\geqc$ are then shown in a second
  table.
  \[
  \begin{array}{l@{\hspace{0.5em}}l}
    \prl{1} & 1\;\geqc\; \D(1,1).\\
    \prl{2} & \D(1,\D(1,\D(1,1)))\; \geqc\; \D(\D(1,\D(1,1)),1).\\
    \prl{3} & \D(1,\D(1,\D(1,1)))\; \gtrc\; \D(\D(1,1),1).\\
    \prl{4} & \D(1,\D(1,\D(1,\D(1,1))))\; \gtrc\; \D(\D(1,\D(1,1)),\D(1,\D(1,1))).\\
    \prl{5} & \D(1,\D(2,\D(3,3)))\; \gtrc\; \D(4,\D(3,3)).\\
  \end{array}
  \]
  \[
  \begin{array}{l@{\hspace{0.5em}}l@{\hspace{0.5em}}l}
    & \{f \in \DC \mid d \supterm f\} & \{f \in \DC \mid e \supterm f\}\\\midrule
    \pref{1} & \emptyset & \emptyset\\
    \pref{2} & \{\D(1,1),\; \D(1,\D(1,1))\} & \{\D(1,1),\; \D(1,\D(1,1))\}\\
    \pref{3} & \{\D(1,1),\; \D(1,\D(1,1))\} & \{\D(1,1)\}\\
    \pref{4} & \{\D(1,1),\; \D(1,\D(1,1)),\; \D(1,\D(1,\D(1,1)))\} & \{\D(1,1),\; \D(1,\D(1,1))\}\\
    \pref{5} & \{\D(3,3),\; \D(2,\D(3,3))\} & \{\D(3,3)\}\\
  \end{array}
  \]
\end{exampapp}

\medskip

\noindent
The following proposition relates the compaction orderings to the compacted
size of the compared \dterms.
\begin{propapp}
  \label{prop-co-cs}
  For all \dterms $d,e$ it holds that
  \smallskip
  
  \subpropapp{prop-geqc-csize} If $d \in \DC$ and $d \geqc e$, then
  $\csize(d) \geq \csize(e)$.

  \subpropapp{prop-cgt-csize} If $d \gtrc e$, then $\csize(d) > \csize(e)$.
\end{propapp}

\begin{proof}
  \prlReset{prop-geqc}
  (\ref{prop-geqc-csize})
  The precondition $d \geqc e$ expands into $\{f \in \DC \mid
  d \supterm f\} \supseteq \{f \in \DC \mid e \supterm f\}$, which implies
  \[
  \begin{arrayprf}
    \prl{prf-geqc-csize-aux} &
    |\{f \in \DC \mid d \supterm f\}|
    \geq  |\{f \in \DC \mid e \supterm f\}|.
  \end{arrayprf}
  \]
  The proposition can then be shown with the following sequence of equations,
  explained below:
  \[
  \begin{arrayprfeq}
   \prl{prf-geqc-1} && \csize(d)\\
   \prl{prf-geqc-2} & = & |\{f \in \DC \mid d \suptermq f \}|\\
   \prl{prf-geqc-3} & = & 1+ |\{f \in \DC \mid d \supterm f\}| 
  \hspace{1em} \\
  \prl{prf-geqc-4} & \geq & 1+ |\{f \in \DC \mid e \supterm f\}|
  \hspace{1em}\\
  \prl{prf-geqc-5} & \geq & |\{f \in \DC \mid e \suptermq f \}|\\
  \prl{prf-geqc-6} & = & \csize(e).
  \end{arrayprfeq}
  \]
  Step~\pref{prf-geqc-3} follows from the precondition $d \in \DC$,
  step~\pref{prf-geqc-4} from
  \pref{prf-geqc-csize-aux}. Steps~\pref{prf-geqc-2} and~\pref{prf-geqc-6}
  are obtained by expanding or contracting, resp., the definition
  of $\csize$. The remaining steps are easy to see.

  (\ref{prop-cgt-csize}) The precondition $d \gtrc e$ expands into $\{f \in
  \DC \mid d \supterm f\} \supset \{f \in \DC \mid e \supterm f\}$, which
  implies
  \[
  \begin{arrayprf}
    \prl{prf-cgt-csize-aux} &
    |\{f \in \DC \mid d \supterm f\}| > |\{f \in \DC \mid e \supterm f\}|.
  \end{arrayprf}
  \]
  The proposition can then be shown with the sequence of equations in the
  proof of Prop.~\ref{prop-geqc-csize}, altered in the following way:
  Step~\pref{prf-geqc-3} is justified since the precondition $d \gtrc e$
  implies $d \in \DC$. In step~\pref{prf-geqc-4}, the relation $\geq$ is
  replaced by $>$, which is justified by~\pref{prf-cgt-csize-aux}.  \qed
\end{proof}

\noindent
The converse statements of Prop.~\ref{prop-geqc-csize}
and~\ref{prop-cgt-csize} do not hold, as demonstrated by
the following example.
\begin{exampapp}
  \label{examp-cocs}
  \prlReset{examp-cocs} The following table shows some counterexamples for the
  converse statements of Prop.~\ref{prop-geqc-csize} and~\ref{prop-cgt-csize},
  that is, \dterms $d$ and $e$ such that $\csize(d) > \csize(e)$ and $d
  \not\geqc e$. As in Example~\ref{examp-co}, the respective values of $\{f
  \in \DC \mid d \supterm f\}$ and $\{f \in \DC \mid e \supterm f\}$ according
  to the definition of $\geqc$ are then shown in a second table.
  \[
  \begin{array}{l@{\hspace{0.5em}}l}
    \prl{1} & \D(1,\D(1,\D(1,\D(1,1))))\; \not \geqc\; \D(1,\D(\D(1,1),1)).\\
    \prl{2} & \D(1,\D(2,\D(3,3)))\; \not \geqc\; \D(4,\D(5,5)).\\
  \end{array}
  \]
  \[
  \begin{array}{l@{\hspace{0.5em}}l@{\hspace{0.5em}}l}
    & \{f \in \DC \mid d \supterm f\} & \{f \in \DC \mid e \supterm f\}\\\midrule
    \pref{1} & \{\D(1,1),\; \D(1,\D(1,1)),\; \D(1,\D(1,\D(1,1)))\}
             & \{\D(1,1),\; \D(\D(1,1),1)\}\\
    \pref{2} & \{\D(3,3),\; \D(2,\D(3,3))\} & \{\D(5,5)\}\\
  \end{array}
  \]
\end{exampapp}

\medskip

\noindent
Theorem~\ref{thm-replace-csize-scsize} can be proven as follows.

\begin{thmref}{}
  {\ref{thm-replace-csize-scsize}}
  Let $d,d',e,e'$ be \mbox{\dterms} such that $e$ occurs in $d$, and $d' =
  \replace{d}{e}{e'}$. It holds that

  \smallskip
  
  \subthmref{thm-replace-csize} If $e \in \DC$ and $e \geqc e'$, then $\csize(d)
  \geq \csize(d')$.

  \subthmref{thm-replace-scsize} If $e \gtrc e'$, then $\scsize(d) >
  \scsize(d')$, where, for all \dterms~$d$ \[\scsize(d) \eqdef \sum_{d
    \suptermq e} \csize(e).\]
\end{thmref}
\begin{proof}
  \prlReset{thm-replace-csize-scsize} We begin with shared aspects of the
  proofs of both subtheorems.  The \dterm $e$ must be in $\DC$, which is
  explicitly stated as precondition for Theorem~\ref{thm-replace-csize} and
  implied by the precondition $e \gtrc e'$ of
  Theorem~\ref{thm-replace-scsize}.  There must exist a set
  $\{d_1,\ldots,d_n\} \subseteq \DC$ for some $n \geq 0$ such that the set $S
  \xeqdef \{f \in \DC \mid d \suptermq f\}$ of compound subterms of $d$ can be
  characterized as the disjoint union of three particular subsets:
  \[
  \begin{arrayprf}
    \prl{prf-thm-rcs-s-disjoint-union} &
    S = \{e\} \uplus \{f \in \DC \mid e \supterm f\} \uplus \{d_1, \ldots,
    d_n\}.
  \end{arrayprf}
  \]
  Let $T$ be the set of those proper subterms of $e$ that are compound and have
  in $d$ an occurrence in a position other than as subterm of $e$. Clearly
  $\{f \in \DC \mid e \supterm f\} \supseteq T$. Thus,
  by~\pref{prf-thm-rcs-s-disjoint-union}) we can characterize $S$ also as
  \[
  \begin{arrayprf}
    \prl{prf-thm-rcs-s-t} &
    S = \{e\} \cup \{f \in \DC \mid e \supterm f\} \cup T \cup \{d_1, \ldots,
    d_n\}.
  \end{arrayprf}
  \]
  The set $S' \xeqdef \{f \in \DC \mid d' \suptermq f\}$ of compound subterms
  of $d'$ can then be characterized as follows:
  \[
  \begin{arrayprf}
    \prl{prf-thm-rcs-sprime} &
    S' = (\{e'\} \cap \DC) \cup \{f \in \DC \mid e' \supterm f\} \cup T\; \cup\\
    & \hphantom{S' =\;} (\{\replace{d_1}{e}{e'}, \ldots, \replace{d_n}{e}{e'}\} \cap \DC).
  \end{arrayprf}
  \]
  From $e \geqc e'$, which is a precondition of
  Theorem~\ref{thm-replace-csize} as well as Theorem~\ref{thm-replace-scsize},
  it follows that $\{f \in \DC \mid e \supterm f\} \supseteq \{f \in \DC \mid
  e' \supterm f\}$. Since $\{f \in \DC \mid e \supterm f\} \supseteq T$ we can
  conclude from~\pref{prf-thm-rcs-sprime} that
  \[
  \begin{arrayprf}
    \prl{prf-thm-rcs-sprime-super} &
    (\{e'\} \cap \DC) \cup \{f \in \DC \mid e \supterm f\} \cup
    \{\replace{d_1}{e}{e'}, \ldots, \replace{d_n}{e}{e'}\}\; \supseteq\; S'.
  \end{arrayprf}
  \]
  We now turn to the two individual subtheorems.

  \smallskip
  
  (\ref{thm-replace-csize}) Since $\csize(d) = |S|$ and $\csize(d') = |S'|$ we
  have to show that $|S| \geq |S'|$.  From~\pref{prf-thm-rcs-sprime-super} it
  follows that $1 + |\{f \in \DC \mid e \supterm f\}| +
  |\{\replace{d_1}{e}{e'}, \ldots, \replace{d_n}{e}{e'}\}| \;\geq\; |S'|$.
  Since clearly $n \geq |\{\replace{d_1}{e}{e'}, \ldots,
  \replace{d_n}{e}{e'}\}|$ it follows that $1 + |\{f \in \DC \mid e \supterm
  f\}| + n \;\geq\; |S'|$.  Since~\pref{prf-thm-rcs-s-disjoint-union} implies
  $|S| = 1 + |\{f \in \DC \mid e \supterm f\}| + n$, that is, $|S|$ can be
  characterized as the left side of the previous disequation, it follows that
  $|S| \geq |S'|$, which concludes the proof of the subtheorem.

  \smallskip
  
  (\ref{thm-replace-scsize}) From~\pref{prf-thm-rcs-sprime-super} it follows
  that
  \[
  \begin{arrayprf}
    \prl{rs-1} &
    \csize(e') + \displaystyle\sum_{e \supterm f} \csize(f) + \sum_{i=1}^n
    \csize(\replace{d_i}{e}{e'})\; \geq\; \scsize(d').
  \end{arrayprf}
  \]
  Given the precondition $e \gtrc e'$ we can conclude by
  Theorem~\ref{thm-replace-csize} that for each $i \in \{1,\ldots,n\}$ it
  holds that $\csize(d_i) \geq \csize(\replace{d_i}{e}{e'})$. Hence:
  \[
  \begin{arrayprf}
    \prl{prf-thm-rcs-si} &
    \displaystyle\sum_{i=1}^n \csize(d_i)\; \geq\; \sum_{i=1}^n
    \csize(\replace{d_i}{e}{e'}).
  \end{arrayprf}
  \]
  From the precondition $e \gtrc e'$ and Prop.~\ref{prop-cgt-csize} it follows
  that $\csize(e) > \csize(e')$. From~\pref{prf-thm-rcs-sprime-super}
  and~\pref{prf-thm-rcs-si} we can then conclude:
  \[
  \begin{arrayprf}
    \prl{prf-thm-rscs-scs-gt} &
    \csize(e) + \displaystyle\sum_{e \supterm f} \csize(f) + \sum_{i=1}^n \csize(d_i)\; >\;
    \scsize(d').
  \end{arrayprf}
  \]
  By~\pref{prf-thm-rcs-s-disjoint-union}, $\scsize(d)$ can be characterized
  as follows:
  \[
  \begin{arrayprf}
    \prl{prf-thm-rscs-scs-s} &
    \scsize(d)\; =\; \csize(e) + \displaystyle\sum_{e \supterm f} \csize(f) + \sum_{i=1}^n
    \csize(d_i).
  \end{arrayprf}
  \]
  Since the right side of~\pref{prf-thm-rscs-scs-s} is identical to left side
  of~\pref{prf-thm-rscs-scs-gt} it follows that $\scsize(d) > \scsize(d')$,
  the conclusion of the subtheorem to be shown.
  \qed
\end{proof}

\noindent
The following example illustrates the \Dterm size measure $\scsize(d)$, which
was defined with Theorem~\ref{thm-replace-scsize}.
\begin{exampapp}
  \
  
  \subexampapp{examp-sc-size-1}
  Let \[d\; \eeqdef\; \D(\D(\D(1,1),\D(1,1)),\D(\D(1,1),1)).\]
  Then the set $\{e \mid d \suptermq e \}$ of
  subterms of~$d$ is
  \[\{1,\;
  \D(1,1),\;
  \D(\D(1,1),1),\;
  \D(\D(1,1),\D(1,1)),\;
  \D(\D(\D(1,1),\D(1,1)),\D(\D(1,1),1))\},\]
  and $\scsize(d) = 0+1+2+2+4 = 9$.

  \smallskip
  
  \subexampapp{examp-sc-size-c}
  If $d,e$ are \dterms such that $\csize(d) > \csize(e)$, then
  it does not necessarily hold that $\scsize(d) \geq \scsize(e)$.
  The following $\dterms$ provide an example:
  \[
  \begin{array}{l@{\hspace{0.5em}}c@{\hspace{0.5em}}l}
    d & \eeqdef & \D(\D(\D(\D(\D(1,1),1),1),1),\D(1,\D(1,\D(1,\D(1,1))))).\\
    e & \eeqdef & \D(\D(\D(\D(\D(\D(\D(1,1),1),1),1),1),1),1).
  \end{array}
  \]
  It holds that $\csize(d) = 8 > 7 = \csize(e)$ but $\scsize(d) = 27 \not \geq
  28 = \scsize(e)$. The calculations of these values are based on the sets
  of subterms of~$d$ and of~$e$, shown in the following, where the compacted
  size of the respective member is annotated at the right:
  \[\begin{array}{l@{\hspace{0.5em}}c@{\hspace{0.5em}}ll@{\hspace{0.5em}}c}
  &&&& \csize\\\midrule
  \{f \mid d \suptermq f \} & = & \{ & 1, & 0\\
  &&& \D(1,1), & 1\\
  &&& \D(1,\D(1,1)), & 2\\
  &&& \D(\D(1,1),1), & 2\\
  &&& \D(1,\D(1,\D(1,1))), & 3\\
  &&& \D(\D(\D(1,1),1),1), & 3\\
  &&& \D(1,\D(1,\D(1,\D(1,1)))), & 4\\
  &&& \D(\D(\D(\D(1,1),1),1),1), & 4\\
  &&& \D(\D(\D(\D(\D(1,1),1),1),1),\D(1,\D(1,\D(1,\D(1,1)))))\; \}. & 8\\[1ex]
  \{f \mid e \suptermq f \} & = & \{ & 1, & 0\\
  &&& \D(1,1), & 1\\
  &&& \D(\D(1,1),1), & 2\\
  &&& \D(\D(\D(1,1),1),1), & 3\\
  &&& \D(\D(\D(\D(1,1),1),1),1), & 4\\
  &&& \D(\D(\D(\D(\D(1,1),1),1),1),1), & 5\\
  &&& \D(\D(\D(\D(\D(\D(1,1),1),1),1),1),1), & 6\\
  &&& \D(\D(\D(\D(\D(\D(\D(1,1),1),1),1),1),1),1)\; \}. & 7
  \end{array}
  \]
  Hence $\csize(d) = 8$, $\scsize(d) = 0+1+2+2+3+3+4+4+8 = 27$,
  $\csize(e) = 7$ and $\scsize(e) = 0+1+2+3+4+5+6+7 = 28$.
\end{exampapp}  

\medskip
\noindent
Two particular characteristics of subproof replacements
according to Theorem~\ref{thm-replace-csize-scsize} are
demonstrated with the following example.

\pagebreak
\begin{exampapp}
  \label{examp-replace}
  \ \par
  \subexampapp{examp-replace-strict} This example is a case that shows that
  strengthening the precondition $e \geqc e'$ of
  Theorem~\ref{thm-replace-csize} to $e \gtrc e'$ does not permit the stronger
  conclusion $\csize(d) > \csize(d')$.  Let
  \[\begin{array}{l@{\hspace{0.5em}}c@{\hspace{0.5em}}l}
  d & \eeqdef & \D(\D(1,\D(1,1)),\D(1,\D(1,\D(1,1)))).\\
  d' & \eeqdef & \D(\D(1,\D(1,1)),\D(\D(1,1),1)).\\
  e & \eeqdef & \D(1,\D(1,\D(1,1))).\\
  e' & \eeqdef & \D(\D(1,1),1).
  \end{array}
  \]
  Then~$e$ occurs in~$d$ and $d' = \replace{d}{e}{e'}$, matching the
  preconditions of Theorem~\ref{thm-replace-csize-scsize}.  Moreover, it holds
  that $e \gtrc e'$. By Theorem~\ref{thm-replace-csize} it follows that
  $\csize(d) \geq \csize(d')$. Indeed, $\csize(d) = \csize(d') = 4$.  By
  Theorem~\ref{thm-replace-scsize} it follows that $\scsize(d) >
  \scsize(d')$. Indeed, $\scsize(d) = 10$ and $\scsize(d') = 9$.  These
  properties and values can be determined on the basis of the following
  intermediate results: That $e > e'$ follows since
  \[\{f \in D \mid e \supterm f \} =
  \{\D(1,1),\; \D(1,\D(1,1))\} \supset \{\D(1,1)\} = \{f \in D \mid e'
  \supterm f \}.\] The sets $\{f \mid d \suptermq f \}$ and $\{f \mid d'
  \suptermq f\}$ underlying the calculation of $\csize(d)$, $\scsize(d)$,
  $\csize(d')$ and $\scsize(d')$ are as follows, where the compacted size of
  the respective member is annotated at the right:
  \[\begin{array}{l@{\hspace{0.5em}}c@{\hspace{0.5em}}ll@{\hspace{0.5em}}c}
  &&&& \csize\\\midrule
  \{f \mid d \suptermq f \} & = & \{ & 1, & 0\\
  &&& \D(1,1), & 1\\
  &&& \D(1,\D(1,1)), & 2\\
  &&& \D(1,\D(1,\D(1,1))), & 3\\
  &&& \D(\D(1,\D(1,1)),\D(1,\D(1,\D(1,1))))\;\}. & 4\\[1ex]
  \{f \mid d' \suptermq f \} & = & \{ & 1, & 0\\
  &&& \D(1,1), & 1\\
  &&& \D(1,\D(1,1)), & 2\\
  &&& \D(\D(1,1),1), & 2\\
  &&& \D(\D(1,\D(1,1)),\D(\D(1,1),1))\; \}. & 4
  \end{array}
  \]

  \subexampapp{examp-replace-twice} This example illustrates that the
  simultaneous replacement of \emph{all} occurrences of $e$ in $d$ by $e'$ is
  essential for Theorem~\ref{thm-replace-csize-scsize} and that $d'$, the
  formula after the replacement, can contain occurrences of~$e$ again. Let
  \[\begin{array}{l@{\hspace{0.5em}}c@{\hspace{0.5em}}l}
  d & \eeqdef &  \D(\D(\D(1,\D(1,\D(1,1))),1),\D(\D(1,\D(1,\D(1,1))),1)).\\
  d' & \eeqdef & \D(\D(\D(1,\D(1,1)),1),\D(\D(1,\D(1,1)),1)).\\
  d'' & \eeqdef & \D(\D(\D(1,\D(1,1)),1),\D(\D(1,\D(1,\D(1,1))),1)).\\
  e & \eeqdef & \D(1,\D(1,1)).\\
  e' & \eeqdef & \D(1,1).
  \end{array}
  \]
  Then~$e$ occurs in~$d$ and $d' = \replace{d}{e}{e'}$, matching the
  preconditions of Theorem~\ref{thm-replace-csize-scsize}.  Moreover, it holds
  that $e \gtrc e'$. By Theorem~\ref{thm-replace-csize} it follows that
  $\csize(d) \geq \csize(d')$. Indeed, $\csize(d) = 5$ and $\csize(d') = 4$.
  Notice that $e$ occurs in $d'$, actually twice. The \dterm~$d''$ is obtained
  from $d$ by replacing just a single occurrence of $e$ with $e'$. Its
  compacted size is $\csize(d'') = 6$, thus not less than or equal to that
  of~$d$. The sets of compound subterms of $d$, $d'$, and $d''$ which underlie
  the determination of their compacted size are as follows:
  \[\begin{array}{l@{\hspace{0.5em}}c@{\hspace{0.5em}}ll}
  \{f \in \DC \mid d \suptermq f \} & = & \{
  & \D(1,1),\\
  &&& \D(1,\D(1,1)),\\
  &&& \D(1,\D(1,\D(1,1))),\\
  &&& \D(\D(1,\D(1,\D(1,1))),1),\\
  &&& \D(\D(\D(1,\D(1,\D(1,1))),1),\D(\D(1,\D(1,\D(1,1))),1))\;\}.\\[1ex]
  \{f \in \DC \mid d' \suptermq f \} & = & \{
  & \D(1,1),\\
  &&& \D(1,\D(1,1)),\\
  &&& \D(\D(1,\D(1,1)),1),\\
  &&& \D(\D(\D(1,\D(1,1)),1),\D(\D(1,\D(1,1)),1))\;\}.\\[1ex]
  \{f \in \DC \mid d'' \suptermq f \} & = & \{
  & \D(1,1),\\
  &&& \D(1,\D(1,1)),\\
  &&& \D(1,\D(1,\D(1,1))),\\
  &&& \D(\D(1,\D(1,1)),1),\\
  &&& \D(\D(1,\D(1,\D(1,1))),1),\\
  &&& \D(\D(\D(1,\D(1,1)),1),\D(\D(1,\D(1,\D(1,1))),1))\;\}.
  \end{array}
  \]
\end{exampapp}  

\medskip

\noindent
Proposition~\ref{prop-co-number} can be proven as follows.

\begin{propref}{}
  {\ref{prop-co-number}}
  For all \dterms $d$ it holds that
  \[\begin{array}{r@{\hspace{0.5em}}l}
  & |\{e \mid d \geqc e \text{ and } \dprim(e) \subseteq \dprim(d)\}|\\
  = & (\csize(d) - 1 + |\dprim(d)|)^2 + |\dprim(d)|.
  \end{array}
  \]
\end{propref}

\begin{proof}
  Let $S$ be the set whose cardinality is denoted by the left side of the
  proposition. Then
  \[\begin{arrayprfeq}
  \prl{1} && S\\
  \prl{2} & =  & \{e \mid d \geqc e \text{ and } \dprim(e) \subseteq
  \dprim(d)\}\\
  \prl{3} & = & \{e \mid
  \{f \in \DC \mid d \supterm f\} \supseteq
  \{f \in \DC \mid e \supterm f\} \text{ and } \dprim(e) \subseteq
  \dprim(d)\}\\
  \prl{4} & =  & \{\D(d_1,d_2) \mid d \supterm d_1 \text{ and } d \supterm d_2\}
  \uplus \dprim(d)\}.
  \end{arrayprfeq}
  \]
  Since $\{e \mid d \supterm e\} = \{e \in \DC \mid d \supterm e\} \uplus
  \dprim(d)$ and $\csize(d)$ is defined as $|\{e \in \DC \mid d \suptermq e
  \}|$ it follows that
  \[\begin{arrayprfeq}
  \prl{5} && |\{e \mid d \supterm e\}| = \csize(d) - 1 + |\dprim(d)|.
  \end{arrayprfeq}
  \]
  From the representation of $S$ in the form~\pref{4} and~\pref{5} it
  follows that $|S| = (\csize(d) - 1 + |\dprim(d)|)^2 +  |\dprim(d)|$,
  that is, the proposition statement.
  \qed
\end{proof}

We now prepare the proofs of Theorems~\ref{thm-sr-ipt} and~\ref{thm-sr-mgt}.
The following proposition shows a specific way to pass between sets of pairs
of terms and most general unifiers.
\begin{propapp}[{\cite[Lemma~4.6]{eder:subst:1985}}]
  If $M, N$ are sets of pairs of terms and $\sigma$ is a most general unifier
  of $M$, then
  
  \subpropapp{prop-mgu-compose-unifiable}
  $M \cup N$ is unifiable if and only if $N\sigma$ is unifiable.
  
  \subpropapp{prop-mgu-compose-asymmetric} If $\tau$ is a most general unifier of
  $N\sigma$, then $\sigma\tau$ is a most general unifier of $M \cup N$.
\end{propapp}
Theorems~\ref{thm-sr-ipt} and~\ref{thm-sr-mgt} both are straightforward
consequences of on an underlying property that is stated below as
Lemma~\ref{lem-sr-mult}. The proof of that lemma involves several applications
of the following further lemma about a way to decompose unifiers associated
with a \DTerm.
\begin{lemapp}
  \label{lem-dds-mult}
  Let $d$ be a \dterm and let $p_1, \ldots, p_n, q$, where $n \geq 0$, be
  positions in $\pos(d)$ such that for all $i \in \{1,\ldots,n\}$ it holds
  that $p_i \not < q$.  Then
  \[\vy_{q}\sigma\; \variant\;
  \vy_{q}\gamma\mgu(\{\{\vy_{p_1}\gamma,\, \vy_{p_1}\tau\gamma\},
  \ldots, \{\{\vy_{p_n}\gamma,\, \vy_{p_n}\tau\gamma\}\}),\] where
    \[\begin{array}{l@{\hspace{0.5em}}c@{\hspace{0.5em}}l@{\hspace{0.5em}}r@{\hspace{0.5em}}c@{\hspace{0.5em}}l}
  \sigma & = & \mgu(\{\pairing_{\daxioms}(d, r) \mid r \in \pos(d)\}),\\
  \tau & = & \mgu(\{\pairing_{\daxioms}(d, r) \mid r \in \pos(d)
  \text{ and } p_i \leq r \text{ for some } i \in \{1,\ldots,n\}\}), \text{ and}\\
  \gamma & = & \mgu(\{\pairing_{\daxioms}(d, r) \mid r \in \pos(d)
  \text{ and } p_i \not \leq r \text{ for all } i \in \{1,\ldots,n\}\}).\\
  \end{array}
    \]
\end{lemapp}

\begin{proof}
  \prlReset{lem-dds-mult-new}
  Let
  \[\begin{array}{l@{\hspace{0.5em}}c@{\hspace{0.5em}}l@{\hspace{0.5em}}r@{\hspace{0.5em}}c@{\hspace{0.5em}}l}
  S & \xeqdef & \{\pairing_{\daxioms}(d, r) \mid r \in \pos(d)\},\\
  T & \xeqdef & \{\pairing_{\daxioms}(d, r) \mid r \in \pos(d)
  \text{ and } p_i \leq r \text{ for some } i \in \{1,\ldots,n\}\},\\
  G & \xeqdef & \{\pairing_{\daxioms}(d, r) \mid r \in \pos(d)
  \text{ and } p_i \not \leq r \text{ for all } i \in \{1,\ldots,n\}\}.
  \end{array}\]
  Then $\sigma = \mgu(S)$, $\tau = \mgu(T)$, and $\gamma = \mgu(G)$.  From the
  definition of $\pairing$ (Def.~\ref{def-pairing}) and the precondition $p_i
  \not < q$ for all $i \in \{1,\ldots,n\}$ it follows that:
  \[
  \begin{arrayprf}
    \prl{varst} & \vars(T) \subseteq \{\posvar_{r} \mid
    p_i \leq r \text{ for some } i \in \{1,\ldots,n\}\}.\\
    \prl{varsg} & \vars(G) \subseteq \{\posvar_{r} \mid
    p_i \not\leq r \text{ for all } i \in \{1,\ldots,n\}\}
    \cup \{y_{p_1}, \ldots, y_{p_n}\}.\\
    \prl{varsprime} & y_{q} \in \{\posvar_{r} \mid
    p_i \not\leq r \text{ for all } i \in \{1,\ldots,n\}\}
    \cup \{y_{p_1}, \ldots, y_{p_n}\}.
  \end{arrayprf}
  \]
  The lemma can now be shown in the following steps, explained below:
  \[
  \begin{arrayprfeq}
  \prl{1} & & \vy_{q}\sigma\\      
  \prl{2}  & = & \vy_{q}\mgu(S)\\  
  \prl{3}  & = & \vy_{q}\mgu(T \cup G)\\
  \prl{4}  & \variant & \vy_{q}\tau\mgu(G\tau)\\
  \prl{5}  & = & \vy_{q}\tau|_{\{\vy_{p_1},\ldots,\vy_{p_n}\}}\mgu(G\tau|_{\{\vy_{p_1},\ldots,\vy_{p_n}\}})\\
  \prl{6}  & \variant & \vy_{q}\mgu(\{\{\vy_p,\vy_p\tau\}\} \cup G)\\
  \prl{7}  & \variant &
  \vy_{q}\gamma\mgu(\{\{\vy_p\gamma,\vy_p\tau\gamma\}\}).\\
  \end{arrayprfeq}
  \]
  Step~\pref{2} is obtained by expanding the definition of $\sigma$, and
  step~\pref{3} follows since $S = T \cup G$.  Step~\pref{4} is obtained by
  Prop.~\ref{prop-mgu-compose-asymmetric}.  By~\pref{varsg} and~\pref{varst}
  it follows that $\vars(G) \cap \vars(T) \subseteq
  \{\vy_{p_1},\ldots,\vy_{p_n}\}$ and by~\pref{varsprime} and~\pref{varst}
  that $\{\vy_{q}\} \cap \vars(T) \subseteq
  \{\vy_{p_1},\ldots,\vy_{p_n}\}$. Since $\dom(\tau) \subseteq \vars(T)$ we
  can replace $\tau$ in~\pref{4} with its restriction to
  $\{\vy_{p_1},\ldots,\vy_{p_n}\}$ and obtain~\pref{5}.  Step~\pref{6} follows
  from Prop.~\ref{prop-mgu-compose-asymmetric} since $\tau|_{\{\vy_{p_1},
    \ldots, \vy_{p_n}\}} \variant \mgu(\{\{\vy_{p_1}, \vy_{p_1}\tau\},\ldots,
  \{\vy_{p_n}, \vy_{p_n}\tau\}\})$.  Finally, step~\pref{7} is obtained by
  Prop.~\ref{prop-mgu-compose-asymmetric} and the definition of~$\gamma$.\qed
\end{proof}
We are now ready to prove the core lemma that shows how the subsumption
relationship between replaced subterm occurrences and a replacing \dterm
transfers to the subsumption relationship between the containing \dterms,
before and after the replacement. It is the basis of Theorems~\ref{thm-sr-ipt}
and~\ref{thm-sr-mgt} below, which express practically useful conditions for
subterm replacement of \dterms. The setting of the lemma is illustrated in
Fig.~\ref{fig-lem-sr-mult}.
\begin{figure}[t]
  \centering
  \scalebox{0.8}{
\begin{tikzpicture}
  \tikzset{
    dot/.style = {circle, fill, minimum size=13pt,
      inner sep=0pt, outer sep=0pt, draw=black},
  }
  \draw (0,0) {} -- (5.5,0) -- (2.75,4.76) -- cycle;
  \fill (0.77,0) {} -- (0.77+1.83,0) -- (0.77+0.91,1.58) -- cycle;  
  \fill (3.36,0) {} -- (3.36+1.375,0) -- (3.36+0.69,1.19) -- cycle;

  \node [dot,fill=white] at (2.75,4.76) {$\emptypos$};
  \node [dot,fill=black!10] at (0.77+0.91,1.58) {$p_1$};
  \node [dot,fill=black!10] at (3.36+0.69,1.19) {$p_2$};
  \node [dot,fill=white] at (2.38,2.38) {$q$};
\end{tikzpicture}
\hspace{0.5cm}
\begin{tikzpicture}
  \tikzset{
    dot/.style = {circle, fill, minimum size=13pt,
      inner sep=0pt, outer sep=0pt, draw=black},
  }

  \fill [black!20] (0.77+0.91,1.58) {} -- (0.77+0.91-0.55,1.58-0.95) --
  (0.77+0.91+0.55,1.58-0.95) -- cycle;

  \fill [black!20] (3.36+0.69,1.19) {} -- (3.36+0.69-0.55,1.19-0.95) --
  (3.36+0.69+0.55,1.19-0.95) -- cycle;

  \draw (0,0) {} -- (0.77,0) -- (0.77+0.91,1.58)
  -- (0.77+1.83,0) -- (3.36,0) -- (3.36+0.69,1.19)
  -- (3.36+1.375,0) -- (5.5,0) -- (2.75,4.76) -- cycle;

  \node [dot,fill=white] at (2.75,4.76) {$\emptypos$};
  \node [dot,fill=black!10] at (0.77+0.91,1.58) {$p_1$};
  \node [dot,fill=black!10] at (3.36+0.69,1.19) {$p_2$};
  \node [dot,fill=white] at (2.38,2.38) {$q$};

  \node [] at (0.77+0.91,1.58-0.6) {$e$};
  \node [] at (3.36+0.69,1.19-0.6) {$e$};
  
\end{tikzpicture}
}
\caption{The setting of Lemma~\ref{lem-sr-mult} for $n=2$.  The left side
  illustrates the \dterm~$d$. Of the positions~$p_1$ and~$p_2$ neither one is
  below the other one.  Position~$q$ must be neither strictly below $p_1$ nor
  strictly below $p_2$. That is, $q$ can be anywhere in the white area
  including $\emptypos$, or it can be one of $p_1$ or $p_2$.  The right side
  illustrates the \dterm $d[e]_{p_1}[e]_{p_2}$ which is obtained from~$d$ by
  replacing the subterms at~$p_1$ and~$p_2$ with occurrences of the
  \dterm~$e$, indicated by smaller gray triangles.}
\label{fig-lem-sr-mult}
\vspace{-20pt}
\end{figure}
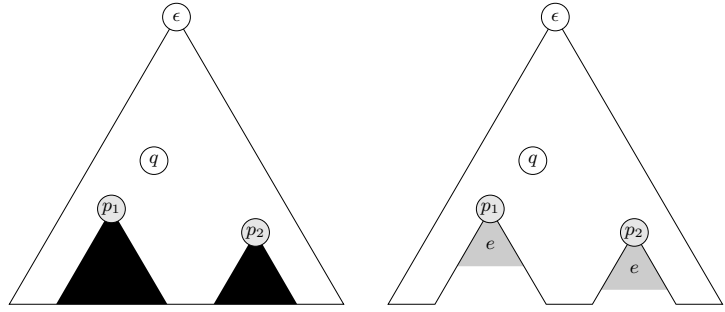
\begin{lemapp}
  \label{lem-sr-mult}
  Let $d, e$ be \dterms, let~$\daxioms$ be an axiom assignment for $d$ and for
  $e$, and let $p_1, \ldots, p_n, q$, where $n \geq 0$, be positions in
  $\pos(d)$ such that for all $i,j \in \{1,\ldots,n\}$ with $i \neq j$ it
  holds that $p_i \not \leq p_j$ and for all $i \in \{1,\ldots,n\}$ it holds
  that $p_i \not < q$.  If for all $i \in \{1, \ldots, n\}$ it holds that
  \[\ipt_{\daxioms}(d, p_i) \subsumedBy \mgt_{\daxioms}(e),\] then
  \[\ipt_{\daxioms}(d,q) \subsumedBy
  \ipt_{\daxioms}(d[e]_{p_1}[e]_{p_2}\ldots [e]_{p_n}, q).\]
\end{lemapp}
\begin{proof}
  \prlReset{lem-sr-mult}
  Define the shorthand $d' = d[e]_{p_1}[e]_{p_2}\ldots [e]_{p_n}$. That is,
  $d'$ is $d$ with the subterm occurrences at $p_1, \ldots, p_n$ replaced
  by~$e$. Define the following sets of pairs of terms and substitutions:
  \[\begin{array}{l@{\hspace{0.5em}}c@{\hspace{0.5em}}l@{\hspace{0.5em}}r@{\hspace{0.5em}}c@{\hspace{0.5em}}l}
  S &  \xeqdef & \{\pairing_{\daxioms}(d, r) \mid r \in \pos(d)\}.\\
  T &  \xeqdef & \{\pairing_{\daxioms}(d, r) \mid r \in \pos(d)
  \text{ and } p_i \leq r \text{ for some } i \in \{1,\ldots,n\}\}.\\
  T' &  \xeqdef & \{\pairing_{\daxioms}(d', r) \mid r \in \pos(d')
  \text{ and } p_i \leq r \text{ for some } i \in \{1,\ldots,n\}\}.\\
  G &  \xeqdef & \{\pairing_{\daxioms}(d, r) \mid r \in \pos(d)
  \text{ and } p_i \not \leq r \text{ for all } i \in \{1,\ldots,n\}\}.\\
  \sigma &  \xeqdef &  \mgu(S).\\
  \tau &  \xeqdef & \mgu(T).\\
  \tau' &  \xeqdef & \mgu(T').\\
  \gamma &   \xeqdef & \mgu(G).\\
  \mu & \xeqdef & \mgu(\{\{\vy_{p_1}\gamma,\vy_{p_1}\ssubold\gamma\},
  \ldots, \{\vy_{p_n}\gamma,\vy_{p_n}\ssubold\gamma\}\}).\\
  \nu & \xeqdef & \mgu(\{\{\vy_{p_1}\gamma,\vy_{p_1}\ssubnew\gamma\},
  \ldots, \{\vy_{p_n}\gamma,\vy_{p_n}\ssubnew\gamma\}\}).
  \end{array}
  \]
  Because the detailed proof is lengthy, we present it modularized into four
  parts, \name{(I)~Conversion of the Preconditions}, \name{(II)~Determining
    the Instantiating Substitution $\rho$}, \textit{(III)~Contexts where
    $\rho$ is Void}, and \textit{(IV)~Deriving the
    Conclusion}. Figure~\ref{fig-lem-sr-mult} may help to get an intuitive
  overview of the parameters of the proven lemma.

    \bigskip
  \noindent{\textit{Part I. Conversion of the Preconditions}}
  \smallskip
  
  \noindent
  The following step is a precondition of the lemma to be proven.
  \[
  \begin{arrayprf}
    \prl{pre-disjoint} &
    p_i \not \leq p_j, \text{ for all } i,j \in \{1,\ldots,n\} \text{ with } i \neq j.
  \end{arrayprf}
  \]  
  The following statements whose proof is described below show that $\sigma$
  when applied to $\vy_q$ and $\vy_{p_i}$ can be decomposed into $\gamma$
  followed by $\mu$.
  \[
  \begin{arrayprf}
    \prl{pi-gamma-mu} & y_{p_i}\sall\; =\; y_{p_i}\gamma\mu,
    \text{ for all } i \in \{1,\ldots,n\}.\\
    \prl{q-gamma-mu} & y_q\sall\; =\; y_q\gamma\mu.\\
  \end{arrayprf}
  \]
  Step~\pref{pi-gamma-mu} follows from Lemma~\ref{lem-dds-mult} with its
  parameters $p_1, \ldots, p_n$ instantiated by the positions of the same name
  in the lemma to be proven but its parameter~$q$ instantiated to $p_i$ for an
  arbitrary $i \in \{1,\ldots,n\}$. The precondition $p_i \not < q$ for all $i
  \in \{1,\ldots,n\}$ of Lemma~\ref{lem-dds-mult} then instantiates to $p_j
  \not < p_i$ for all $j \in \{1,\ldots,n\}$, which follows
  from~\pref{pre-disjoint}.  Step~\pref{q-gamma-mu} follows from
  Lemma~\ref{lem-dds-mult} with all of its parameters $p_1, \ldots, p_n, q$
  instantiated by the positions of the same names in the lemma to be proven.
  
  Let us consider now the precondition $\ipt_{\daxioms}(d, p_i) \subsumedBy
  \mgt_{\daxioms}(e)$ for an arbitrary $i \in \{1,\ldots,n\}$. Its left side
  can be converted by expanding and contracting definitions and
  step~\pref{pi-gamma-mu} as follows:
  \[
  \begin{arrayprfeq}
    \prl{x1} && \ipt_{\daxioms}(d, p_i)\\
    \prl{x2} & = &  \P(y_{p_i}\mgu(\{\pairing_{\daxioms}(d, r) \mid r \in \pos(d)\}))\\
    \prl{x3} & = &  \P(y_{p_i}\mgu(S))\\
    \prl{x4} & = &  \P(y_{p_i}\sigma)\\
    \prl{x5} & = &  \P(y_{p_i}\gamma\mu).\\
  \end{arrayprfeq}
  \]
  The conversion of the right side of the considered precondition is based on
  some auxiliary definitions and statements.  For all $i \in \{1,\ldots,n\}$
  define the following sets of pairs of terms and substitutions:
  \[\begin{array}{l@{\hspace{0.5em}}c@{\hspace{0.5em}}l@{\hspace{0.5em}}r@{\hspace{0.5em}}c@{\hspace{0.5em}}l}
  T'_i & \xeqdef & \{\pairing_{\daxioms}(d', r) \mid r \in \pos(d')
  \text{ and } p_i \leq r\}.\\
  \NT'_i & \xeqdef & \bigcup_{j \in \{1,\ldots,n\}\setminus\{i\}} T'_j.
  \end{array}\]
  Then, as explained below, for all $i,j \in \{1,\ldots,n\}$ the
  following holds:
  \[
  \begin{arrayprf}
    \prl{tpi1} & T'_i \cup \NT'_i = T'.\\
    \prl{tpi2} & \vars(T'_i) \subseteq \posvar(\{r \mid p_i \leq r\}).\\
    \prl{tpi3} & \text{if } i \neq j, \text{ then } \vars(T'_i) \cap \vars(T'_j)
    = \emptyset.\\
    \prl{tpi4} & \vars(\NT'_i) \cap \{y_{p_i}\} = \emptyset.\\
    \prl{tpi5} & \vars(\NT'_i) \cap \vars(T'_i) = \emptyset.\\
    \prl{tpi6} & y_{p_i}\tau' = y_{p_i}\mgu(T'_i).\\
    \prl{idx-ssubnew-dom} &
    \text{If } y_{p_i} \in \dom(\ssubnew), \text{ then }
    \vars(y_{p_i}\ssubnew) \subseteq \posvar(\{r \mid p_i < r\}).\\
    \prl{nonoverlap-term} & \text{If } i \neq j, \text{ then }
    \vars(y_{p_i}\ssubnew) \cap \vars(y_{p_j}\ssubnew) = \emptyset.\\
  \end{arrayprf}
  \]
  Step~\pref{tpi1} follows immediately from the definitions of $T'_i$,
  $\NT'_i$ and $T$.  Step~\pref{tpi2} follows from the definition of $T'_i$
  and the definition of $\pairing$ (Def.~\ref{def-pairing}). Step~\pref{tpi3}
  follows from~\pref{tpi2} and~\pref{pre-disjoint}.  Step~\pref{tpi4} follows
  from the definition of $\NT'_i$ and steps~\pref{tpi2} and~\pref{pre-disjoint}.
  Step~\pref{tpi5} follows from the definition of $\NT'_i$ and
  step~\pref{tpi3}.  Step~\pref{tpi6} follows from the definition of~$\tau'$
  and steps~\pref{tpi1}, \pref{tpi4} and~\pref{tpi5}.
  Step~\pref{idx-ssubnew-dom} follows from~\pref{tpi6} and~\pref{tpi2}.
  Step~\pref{nonoverlap-term} follows from~\pref{tpi6}, \pref{tpi3} and
  \pref{pre-disjoint}.

  The right side of the precondition $\ipt_{\daxioms}(d, p_i) \subsumedBy
  \mgt_{\daxioms}(e)$ can now be converted in the following steps described
  below:
  \[
  \begin{arrayprfeq}
    \prl{y1} && \mgt_{\daxioms}(e)\\
    \prl{y2} & = & \P(y_{\emptypos}\mgu(\{\pairing_{\daxioms}(e, r) \mid r \in
    \pos(e)\}))\\
    \prl{y3} & \variant
    & \P(y_{\emptypos}\mgu(\{\pairing_{\daxioms}(e, r) \mid r \in \pos(e)\})\sshift{p_i})\\
    \prl{y4} & =
    & \P(y_{\emptypos}\mgu(\{\pairing_{\daxioms}(d'|_{p_i}, r) \mid r \in \pos(d'|_{p_i})\})\sshift{p_i})\\
    \prl{y5} & =
    & \P(y_{p_i}\mgu(\{\pairing_{\daxioms}(d', r) \mid r \in \pos(d') \text{ and } p_i \leq r\}))\\
    \prl{y6} & = & \P(y_{p_i}\mgu(T'_i))\\
    \prl{yx10} & = & \P(y_{p_i}\tau').\\    
  \end{arrayprfeq}
  \]
  Step~\pref{y2} is obtained from~\pref{y1} by expanding the definition of
  $\mgt$.  Step~\pref{y3} follows from Prop.~\ref{prop-shift-variant},
  step~\pref{y4} since by the definition of $d'$ it holds that $d'|_{p_i} = e$,
  and step~\pref{y5} from Prop.~\ref{prop-tree-renaming-add}.  Step~\pref{y6}
  is obtained by contracting the definition of $T'_i$. Step~\pref{yx10}
  follows from~\pref{tpi6}. Note that \pref{y1} is independent from~$i$ and
  the conversion of~\pref{y1} to~\pref{yx10} is possible for any $i \in
  \{1,\ldots,n\}$.
  
  Because~\pref{x1} and~\pref{x5} as well as~\pref{y1} and~\pref{yx10} are
  equal, we can now reformulate the precondition that for all $i \in
  \{1,\ldots,n\}$ it holds that $\ipt_{\daxioms}(d, p_i) \subsumedBy
  \mgt_{\daxioms}(e)$ as
  \[\begin{arrayprf}
  \prl{pre-v1} & y_{p_i}\gamma\mu \subsumedBy y_{p_i}\tau',
  \text{ for all } i \in \{1,\ldots,n\}.
  \end{arrayprf}\]

  \smallskip
  \noindent{\textit{Part II. Determining the Instantiating Substitution $\rho$}}
  \smallskip

  \noindent
  We show, as explained below, that for all $i \in \{1,\ldots,n\}$ there
  exists a substitution~$\rho_i$ with the following properties:
  \[
  \begin{arrayprf}
    \prl{pre-i} & y_{p_i}\gamma\mu\; =\; y_{p_i}\ssubnew\rho_i.\\
    \prl{pre-rho-dom} &
    \dom(\rho_i) \subseteq \vars(y_{p_i}\ssubnew).\\
    \prl{pre-i-dom-2} &
    \text{If } y_{p_i} \in \dom(\ssubnew), \text{ then }
    \dom(\rho_i) \subseteq \posvar(\{r \mid p_i < r\}).\\
    \prl{pre-i-dom-disjoint} &
    \text{If } i \neq j, \text{ then }
    \dom(\rho_i) \cap \dom(\rho_j) = \emptyset.\\
    \prl{pre-i-dom-cap} &
    \dom(\rho_i) \cap \dom(\ssubnew) = \emptyset.\\
  \end{arrayprf}
  \]
  Steps~\pref{pre-i} and~\pref{pre-rho-dom} follow from~\pref{pre-v1}.
  Step~\pref{pre-i-dom-2} follows from~\pref{pre-rho-dom}
  and~\pref{idx-ssubnew-dom}, step~\pref{pre-i-dom-disjoint}
  from~\pref{pre-rho-dom} and~\pref{nonoverlap-term}.
  Step~\pref{pre-i-dom-cap} follows from~\pref{pre-rho-dom} since the
  idempotence of~$\tau'$ is equivalent to $\dom(\tau') \cap \vrng(\tau') =
  \emptyset$, which implies $\vars(\vy_{p_i}\tau') \cap \dom(\tau') =
  \emptyset$.

  Step~\pref{pre-i-dom-disjoint} justifies to define a substitution~$\rho$,
  which combines the substitutions $\rho_i$ by forming their union:
  \[\rho\; \xeqdef\; \bigcup_{i=1}^{n}\{v \mapsto v\rho_i \mid v \in
  \dom(\rho_i)\}.\]
  The substitution~$\rho$ has the following properties, whose derivation
  is described below:
 \[
 \begin{arrayprf}
   \prl{pre-dom-components} & y_{p_i}\ssubnew\rho \; =\;  y_{p_i}\ssubnew\rho_i,
   \text{ for all } i \in \{1,\ldots,n\}.\\
    \prl{pre} & y_{p_i}\gamma\mu\; =\; y_{p_i}\ssubnew\rho,
    \text{ for all } i \in \{1,\ldots,n\}.\\
    \prl{dom-rho-2} & \dom(\rho) \subseteq \posvar(\{r \mid
    p_i \leq r \text{ for some } i \in \{1,\ldots,n\}\}).\\
   \prl{pre-dom-cap} & \dom(\rho) \cap \dom(\ssubnew) = \emptyset.\\
 \end{arrayprf}
  \]
  Step~\pref{pre-dom-components} follows from the definition of~$\rho$, given
  that for all $i,j \in \{1,\ldots,n\}$ with $i \neq j$ it holds that
  $\vars(\vy_{p_i}\tau') \cap \dom(\rho_j) = \emptyset$, which follows from
  \pref{pre-rho-dom} and \pref{nonoverlap-term}.
  Step~\pref{pre} follows from \pref{pre-dom-components} and~\pref{pre-i}.
  Step~\pref{dom-rho-2} follows from the definition of~$\rho$ and
  steps~\pref{pre-rho-dom}, \pref{tpi6}, and~\pref{tpi2}.
  Step~\pref{pre-dom-cap} follows from the definition of~$\rho$ and
  step~\pref{pre-i-dom-cap}.

  \bigskip
  \noindent{\textit{Part III. Contexts where $\rho$ is Void}}
  \smallskip\par
  \noindent
  The variables occurring in members of the range of $\gamma$ as well as
  $\vy_q$ are contained in the same set of position-associated variables:
  \[
  \begin{arrayprf}
    \prl{rng-gamma} & \vrng(\gamma) \subseteq \posvar(\{r \mid p_i \not \leq r
    \text{ for all } i \in \{1,\ldots,n\}\}) \cup \{y_{p_1}, \ldots,
    y_{p_n}\}.\\
    \prl{yq-incl} & y_q \in \posvar(\{r \mid p_i \not \leq r
    \text{ for all } i \in \{1,\ldots,n\}\}) \cup \{y_{p_1}, \ldots,
    y_{p_n}\}.
  \end{arrayprf}
  \]  
  Step~\pref{rng-gamma} follows from the definitions of~$\gamma$ and $G$ and
  the definition of $\pairing$ (Def.~\ref{def-pairing}).  Step \pref{yq-incl}
  follows from the precondition that for all $i \in \{1,\ldots,n\}$ it holds
  that $p_i \not < q$.
  Now, let $y$ be a position related variable and let $v$ be a variable such
  that
  \[\begin{array}{l}
  y \in \{y_{p_1},\ldots,y_{p_n},y_{p_q}\}, \text{ and}\\
  v \in \vars(y\gamma).
  \end{array}\]
  From~\pref{rng-gamma} and~\pref{yq-incl} it follows that
  \[
  \begin{arrayprf}
    \prl{v-in} & v \in \posvar(\{r \mid p_i \not \leq r
    \text{ for all } i \in \{1,\ldots,n\}\}) \cup \{y_{p_1}, \ldots,
    y_{p_n}\}.
  \end{arrayprf}
  \]
  As proven below, then
  \[
  \begin{arrayprf}
    \prl{aux-vy} & v\mu = v\rho\mu.
  \end{arrayprf}
  \]  
  Step~\pref{aux-vy} is proven by considering three cases (the first two
  overlap, the third applies if none of the first two applies):
  \begin{enumerate}
  \item Case $v \notin \{y_{p_1}, \ldots, y_{p_n}\}$.  Then, by~\pref{v-in}
    and~\pref{dom-rho-2}, $v \notin \dom(\rho)$, hence $v\mu = v\rho\mu$.

  \item Case $v \in \dom(\ssubnew)$.  Then, by~\pref{pre-dom-cap}, $v \notin
    \dom(\rho)$, hence $v\mu = v\rho\mu$.
  
  \item Case $v \in \{y_{p_1}, \ldots, y_{p_n}\} \setminus \dom(\ssubnew)$.
    Then, by \pref{pre}, $v\gamma\mu = v\rho$.  Since $v \in \vars(y\gamma)$
    and $\gamma$ is idempotent it follows that $v = v\gamma$. Hence $v\mu =
    v\rho$, and, since $\mu$ is idempotent, $v\mu = v\rho\mu$.
  \end{enumerate}
  Given the definition of $v$ and $y$ we can instantiate~\pref{aux-vy} to the
  following statements about the $y_{p_i} \in \dom(\tau')$ for $i \in
  \{1,\ldots, n\}$ and $y_q$.
  \[
  \begin{arrayprf}
    \prl{gap-1-a} & y_{p_i}\gamma\mu\; =\; y_{p_i}\gamma\rho\mu,
    \text{ for all } i \in \{1,\ldots,n\}.\\
    \prl{gap-3} & \vy_q\gamma\mu = \vy_q\gamma\rho\mu.\\
  \end{arrayprf}
  \]

  \bigskip
  \noindent{\textit{Part IV. Deriving the Conclusion}}
  \smallskip

  \noindent
  The conclusion of the lemma to be proven, that is, \[\ipt_{\daxioms}(d,q)
  \subsumedBy \ipt_{\daxioms}(d[e]_{p_1}[e]_{p_2}\ldots [e]_{p_n}, q)\]
  can be reformulated as
  \[
  \begin{arrayprf}
    \prl{goal} & y_{q}\gamma\mu \subsumedBy y_{q}\gamma\nu.
  \end{arrayprf}
  \]
  For the left side, the reformulation follows since $\ipt_{\daxioms}(d,q) =
  \P(y_{q}\gamma\mu)$, which can be derived analogously to
  steps~\pref{x1}--\pref{x5}, but by applying~\pref{q-gamma-mu} instead
  of~\pref{pi-gamma-mu}. For the right side it follows since
  $\ipt_{\daxioms}(d[e]_{p_1}[e]_{p_2}\ldots [e]_{p_n}, q) =
  \ipt_{\daxioms}(d', q) =\linebreak \P(\vy_q\mgu(\{\pairing_{\alpha}(d',r)
  \mid r \in \pos(d')\})) \variant \P(\vy_q\gamma\nu)$, which can be derived
  by expanding definitions and, for the last step, applying
  Lemma~\ref{lem-dds-mult}.
  
  To prove~\pref{goal}, we need a further auxiliary statement, which is
  derived along with an intermediate step about the domain of~$\gamma$ as
  explained below:
  \[
  \begin{arrayprf}
    \prl{dom-gamma} & \dom(\gamma) \subseteq \posvar(\{r \mid p_i \not \leq r
    \text{ for all } i \in \{1,\ldots,n\}\}) \cup \{y_{p_1}, \ldots,
    y_{p_n}\}.\\
    \prl{gap-2} & y_{p_i}\ssubnew\; =\; 
    y_{p_i}\ssubnew\gamma,
    \text{ for all } i \in \{1,\ldots,n\} \text{ s.th. } y_{p_i} \in \dom(\ssubnew).\\

  \end{arrayprf}
  \]
  Step~\pref{dom-gamma} follows from the definitions of~$\gamma$ and $G$ and
  the definition of $\pairing$ (Def.~\ref{def-pairing}).  Step~\pref{gap-2}
  can be shown as follows: Assume $y_{p_i} \in \dom(\ssubnew)$.
  By~\pref{idx-ssubnew-dom} it follows that $\vars(y_{p_i}\ssubnew) \subseteq
  \posvar(\{r \mid p_i < r\})$. With~\pref{dom-gamma} it follows that
  $\vars(y_{p_i}\ssubnew) \cap \dom(\gamma) = \emptyset$, which
  implies~\pref{gap-2}.  We can now proceed to prove the goal~\pref{goal} as
  follows, explained below.
  \[
  \begin{arrayprf}
    \prl{xunif-00} &  y_{p_i}\gamma\rho\mu\; =\ y_{p_i}\ssubnew\rho,
    \text{ for all } i \in \{1,\ldots,n\}.\\
    \prl{unif-00} &  y_{p_i}\gamma\rho\mu\; =\ y_{p_i}\ssubnew\gamma\rho,
    \text{ for all } i \in \{1,\ldots,n\} \text{ s.th. } y_{p_i} \in \dom(\ssubnew).\\
    \prl{unif-01} & y_{p_i}\gamma\rho\mu\mu\; =\ y_{p_i}\ssubnew\gamma\rho\mu,
    \text{ for all } i \in \{1,\ldots,n\} \text{ s.th. } y_{p_i} \in \dom(\ssubnew).\\
    \prl{unif-1-indom} & y_{p_i}\gamma\rho\mu\;
    =\ y_{p_i}\ssubnew\gamma\rho\mu,
    \text{ for all } i \in \{1,\ldots,n\} \text{ s.th. } y_{p_i} \in
    \dom(\ssubnew).\\
    \prl{unif-1-notindom} & y_{p_i}\gamma\rho\mu\;
    =\ y_{p_i}\ssubnew\gamma\rho\mu,
    \text{ for all } i \in \{1,\ldots,n\}
    \text{ s.th. } y_{p_i} \notin \dom(\ssubnew).\\
    \prl{unif-1} & y_{p_i}\gamma\rho\mu\; =\ y_{p_i}\ssubnew\gamma\rho\mu,
    \text{ for all } i \in \{1,\ldots,n\}.\\
    \prl{unif-2} & \rho\mu \text{ is a unifier of }
    \{\{\vy_{p_1}\gamma,\vy_{p_1}\ssubnew\gamma\},
    \ldots, \{\vy_{p_n}\gamma,\vy_{p_n}\ssubnew\gamma\}\}.\\
    \prl{unif-3} & \rho\mu \subsumedBy \nu.\\
    \prl{goal-0} &  y_q\gamma\rho\mu \subsumedBy y_q\gamma\nu.\\
    \prl{goal-1} &  y_q\gamma\mu \subsumedBy y_q\gamma\nu.\\
  \end{arrayprf}
  \]
  Step~\pref{xunif-00} follows from~\pref{gap-1-a} and~\pref{pre}.
  Step~\pref{unif-00} follows from~\pref{xunif-00} and~\pref{gap-2}.
  Step~\pref{unif-01} follows from \pref{unif-00}.  Step~\pref{unif-1-indom}
  follows from~\pref{unif-01} since~$\mu$ is idempotent.
  Step~\pref{unif-1-notindom} holds since if $y_{p_i} \notin \dom(\ssubnew)$,
  then $y_{p_i}\ssubnew = y_{p_i}$.  Step~\pref{unif-1} follows
  from~\pref{unif-1-indom} and~\pref{unif-1-notindom}.  Step~\pref{unif-2}
  follows from~\pref{unif-1}.  Step~\pref{unif-3} follows from~\pref{unif-2}
  and the definition of~$\nu$.  Step~\pref{goal-0} follows from~\pref{unif-3}.
  Finally, step~\pref{goal-1}, which is the goal to be proven listed above
  as~\pref{goal}, follows from~\pref{goal-0} and~\pref{gap-3}.
  \qed
\end{proof}

\pagebreak
\noindent
Lemma~\ref{lem-sr-mult} is now applied to justify
Theorem~\ref{thm-sr-ipt} and~\ref{thm-sr-mgt}.

\begin{thmref}{}
  {\ref{thm-sr-ipt}}
  Let $d, e$ be \dterms, let~$\daxioms$ be an axiom assignment for $d$ and for
  $e$, and let $p_1, \ldots, p_n$, where $n \geq 0$, be positions in $\pos(d)$
  such that for all $i,j \in \{1,\ldots,n\}$ with $i \neq j$ it holds that
  $p_i \not \leq p_j$.  If for all $i \in \{1, \ldots, n\}$ it holds that
  \[\ipt_{\daxioms}(d, p_i) \subsumedBy \mgt_{\daxioms}(e),\] then
  \[\mgt_{\daxioms}(d) \subsumedBy
  \mgt_{\daxioms}(d[e]_{p_1}[e]_{p_2}\ldots [e]_{p_n}).\]
\end{thmref}
\begin{proof}
  The theorem expresses the special case of Lemma~\ref{lem-sr-mult} with $q =
  \emptypos$. The precondition of that lemma that for all $i \in
  \{1,\ldots,n\}$ it holds that $p_i \not < q$ then holds trivially. The
  remaining preconditions are the same as those of Lemma~\ref{lem-sr-mult}.
  The conclusion is obtained from that of Lemma~\ref{lem-sr-mult} by
  contracting the definition of $\mgt$. \qed
\end{proof}

\begin{thmref}{}
  {\ref{thm-sr-mgt}}
  Let $d, e$ be \dterms and let~$\daxioms$ be an axiom assignment for $d$ and
  for $e$.  For all positions $p \in \pos(d)$ it then holds that if
  \[\mgt_{\daxioms}(d|_p) \subsumedBy \mgt_{\daxioms}(e),\] then
  \[\mgt_{\daxioms}(d) \subsumedBy \mgt_{\daxioms}(d[e]_p).\]
\end{thmref}
\begin{proof}
Follows from Theorem~\ref{thm-sr-ipt} and Prop.~\ref{prop-ipt-subsumedby-mgt}.
\qed
\end{proof}

\subsection{Supplementary Material for Section~\ref{sec-experiments}}

\subsubsection{Notes on the \PRIMES Lemma Computation Method.}

As indicated by column~\xatt{DP} in Table~\ref{tab-bigmer}, subproof~18 is the
largest \dterm in \PMER that is \defname{prime}, as we call \dterms~$d$ such
that $\csize(d) = \tsize(d)$.  The prime property can also be characterized in
three further ways: (1)~Every subterm of $d$ has only a single occurrence in
$d$. (2)~$\height(d) = \tsize(d)$.  (3)~$d$ is in the smallest set~$P$ that
satisfies the following conditions: (i) $1 \in P$. (ii) For all $e \in P$ it
holds that $\d(1,e) \in P$ and $\d(e,1) \in P$.  Subproofs~1--18 are exactly
those in \PMER that are prime.  Moreover, all prime proofs in \PMER are a
subproof of subproof~18. Hence we speak of subproof~18 as ``prime core'' of
\PMER.  Column~\xatt{DS} of Table~\ref{tab-bigmer} indicates a bottom-up
construction of subproof~18 and its subproofs that matches the inductive
characterization~(3) (if the primitive \dterm $\n$ is not distinguished from
$1$).  From the perspective of lemma computation, the objective is then to
find a systematic way in which the set of possible lemmas that are derivable
from the axiom can be narrowed down to a much smaller set that still contains
the prime core.  The number of distinct prime \dterms of a given size~$n$
(tree size or compacted size, which are identical for prime \dterms) grows by
sequence \href{https://oeis.org/A011782}{A011782} of the \OEIS \cite{oeis},
i.e.,~$1$ for $n=0$ and $2^{n-1}$ for $n > 0$, which is much slower than for
\dterms in general, where the growth is with respect to tree size by
\href{https://oeis.org/A000108}{A000108} and to tree size by
\href{https://oeis.org/A254789}{A254789}.  For $n=17$, the size of
subproof~18, there are $2^{16}$ prime \dterms, of which 14,882 have, for \Luk
as axiom, a defined \MGT.  Only two of them, the prime core and another \dterm
with the same \MGT, remain if we, aside of a redundancy criterion (no smaller
prime proof of the \MGT), require that the number of different variables in
the \MGT is the same as in the axiom, i.e., 4. Another possibility that,
however, leads to larger sets can be based on the property that all subproofs
have a weakly organic \MGT.  Of course, the size parameter~17 in the
invocation of \PRIMES has been chosen according to Table~\ref{tab-bigmer}. In
practice, a system could try this form of lemma generation with increasing
values of the size parameter. For axiom \Luk, experiments with other size
values did not lead to a substantial decrease of the compacted size.

\vspace{-12pt}
\subsubsection{Notes on the \PSUB Lemma Computation Method.}
The greatest part of the running times for lemma computation with \PSUB
reported in Table~\ref{tab-exp} was taken for determining the properties
\xatt{RC} (\name{\XC-regular}) and \xatt{TO} (\name{weakly organic}).  The
\MGT of subproof~30 of Table~\ref{tab-bigmer} was for proofs~(7.)--(9.) of
Table~\ref{tab-exp} among the generated lemmas, but reached without passing
through subproof~27, which has $\gtrc$ as value of \xatt{DS}.  \CMProver,
which was used for proofs~(8.) and (9.), is compared to \name{LeanCoP}
\cite{leancop} more like the \name{Prolog Technology Theorem Prover (PTTP)}
\cite{pttp} and \name{SETHEO} \cite{setheo:92} based on a compilation of the
input clauses to Prolog code.  Early experiments where \name{SETHEO} was
combined with bottom-up lemma generation were described in
\cite{schumann:delta:1994}.  In our experiments \CMProver was configured such
that the cost measure underlying iterative deepening is the number of subgoals
\cite{pttp}, reflecting the tree size.

\vspace{-12pt}
\subsubsection{New Short Proofs.}

\enlargethispage{5pt}

Figures~\ref{fig-proof-short32} and~\ref{fig-proof-short191} below show proofs
obtained in the experiments described in Sect.~\ref{sec-experiments}.  The
proofs are shown in Meredith's notation, like Fig.~\ref{fig-representations}c
(p.~\pageref{fig-representations}) and Fig.~\ref{fig-proof-mer}
(p.~\pageref{fig-proof-mer}).  For subproofs whose MGT is also the MGT of a
subproof of~\PMER (Fig.~\ref{fig-proof-mer}) the respective line number in
\PMER is annotated with prefix \textit{M}.  Representations of these proofs as
Prolog-readable \Dterms are provided at
\href{http://cs.christophwernhard.com/cd}{\texttt{http://cs.christophwernhard.com/cd}}.

\begin{figure}
  \vspace{-9pt}
  \begin{tabular}{rR{1.3em}@{\hspace{0.5em}}L{27.1em}@{\hspace{0.0em}}R{3em}@{\hspace{0.5em}}R{3em}}
& 1. & $\g{CCCpqrCCrpCsp}$ & \mer{1}\\
& 2. & $\g{CCCpqpCrp} \mereq \f{D}\f{D}\f{D}1\f{D}111\n$ & \mer{2}\\
& 3. & $\g{CCCpqrCqr} \mereq \f{D}\f{D}\f{D}1\f{D}1\f{D}121\n$ & \mer{3}\\
& 4. & $\g{CpCCpqCrq} \mereq \f{D}31$ & \mer{4}\\
& 5. & $\g{CCCpqCrsCCCqtsCrs} \mereq \f{D}\f{D}\f{D}1\f{D}1\f{D}1\f{D}141\n$ & \mer{5}\\
& 6. & $\g{CCCpqCrsCCpsCrs} \mereq \f{D}51$ & \mer{6}\\
& 7. & $\g{CCCCCpqrsCtpCCrpCtp} \mereq \f{D}\f{D}641$ & \mer{8}\\
& 8. & $\g{CCCCpqCrqCCCqsptCuCCCqspt} \mereq \f{D}17$ & \mer{10}\\
& 9. & $\g{CCCCpqrCsqCCCqtpCsq} \mereq \f{D}5\f{D}\f{D}88\n$ & \mer{12}\\
& 10. & $\g{CCCCpqrsCCsqCpq} \mereq \f{D}96$ & \mer{13}\\
& 11. & $\g{CCpqCCCprqq} \mereq \f{D}7\f{D}\f{D}977$ & \mer{16}\\
& 12. & $\g{CCCCpqrCCCpsqqCCCpsqq} \mereq \f{D}11.11$ &\\
* & 13. & $\g{CCpqCCqrCpr} \mereq \f{D}\f{D}10.12.10$ & \mer{17}\\
* & 14. & $\g{CCCpqpp} \mereq \f{D}12.2$ & \mer{18}\\
* & 15. & $\g{CpCqp} \mereq \f{D}33$ & \mer{19}\\[-5pt]
\end{tabular}
  \caption{A proof of the Tarski-Bernays axioms from \Luk with compacted
    size~32. The subproof of \Syll (i.e., problem~\PSYLL) has compacted
    size~30 and was obtained in experiment~(9.) of Table~\ref{tab-exp}
    (p.~\pageref{tab-exp}).}
  \label{fig-proof-short32}
\end{figure}

\begin{figure}
\begin{tabular}{rR{1.3em}@{\hspace{0.5em}}L{27.1em}@{\hspace{0.0em}}R{3em}@{\hspace{0.5em}}R{3em}}
& 1. & $\g{CCCpqrCCrpCsp}$ & \mer{1}\\
& 2. & $\g{CCCpCCqrCsrCtCruCvCtCru} \mereq \f{D}1\f{D}1\f{D}11$ & \\
& 3. & $\g{CCCpCqrCsCCCtCruvCwvCxCsCCCtCruvCwv} \mereq \f{D}1\f{D}1\f{D}1\f{D}1\f{D}\f{D}\f{D}2111$ & \\
& 4. & $\g{CCCpqCrsCCCqtsCrs} \mereq \f{D}\f{D}\f{D}1\f{D}1\f{D}1\f{D}1\f{D}\f{D}\f{D}\f{D}1\f{D}\f{D}\f{D}121111111$ & \mer{5}\\
& 5. & $\g{CCpCqrCCCpsrCqr} \mereq \f{D}\f{D}\f{D}\f{D}31111$ & \mer{7}\\
& 6. & $\g{CCCCCpqrsCtpCCrpCtp} \mereq \f{D}51$ & \mer{8}\\
& 7. & $\g{CCCCpqCrqCCCqsptCuCCCqspt} \mereq \f{D}16$ & \mer{10}\\
& 8. & $\g{CCCCpqrsCCsqCpq} \mereq \f{D}\f{D}4\f{D}\f{D}771\f{D}41$ & \mer{13}\\
* & 9. & $\g{CCpqCCqrCpr} \mereq \f{D}\f{D}8\f{D}5\f{D}6\f{D}\f{D}\f{D}1\f{D}\f{D}\f{D}\f{D}1\f{D}\f{D}\f{D}1311111118$ & \mer{17}\\[-5pt]
\end{tabular}
\caption{A proof of \Syll from \Luk (i.e., problem~\PSYLL) with tree size~191,
  obtained in experiment~(10.) of Table~\ref{tab-exp} (p.~\pageref{tab-exp}).
  The precomputed table of small proofs involved in computing this proof
  contained 12,090 entries, one prime proof of minimal size for each
  formula that appears as \MGT of a prime \Dterm with size up to 20.}
  \label{fig-proof-short191}
\end{figure}

\end{document}